\documentclass{article}

\PassOptionsToPackage{numbers, compress}{natbib}

    \usepackage[preprint]{neurips_2025}

\usepackage[utf8]{inputenc} 
\usepackage[T1]{fontenc}    
\usepackage{hyperref}       
\usepackage{url}            
\usepackage{booktabs}       
\usepackage{amsfonts}       
\usepackage{graphicx}       
\usepackage{nicefrac}       
\usepackage{microtype}      
\usepackage{xcolor}         
\usepackage{amsmath}
\usepackage{textcomp}
\usepackage{colortbl}
\usepackage{xpatch}
\usepackage{tcolorbox}
\usepackage{mdframed}
\usepackage{enumitem}
\tcbuselibrary{skins}
\usepackage{multirow}
\usepackage{subcaption} 
\makeatletter
\xapptocmd{\NAT@bibsetnum}{\setlength{\leftmargin}{0pt}\setlength{\itemindent}{\labelwidth}\addtolength{\itemindent}{\labelsep}}{}{}
\makeatother

\title{Enhancing Large Language Models through Structured Reasoning}

\author{%
  Yubo Dong \\ 
  Department of Computer Science\\
  Zhejiang University\\
  \texttt{dongyubo@zju.edu.cn} \\
  \And
  Hehe Fan\\
  Department of Computer Science\\
  Zhejiang University\\
  \texttt{hehe.fan.cs@gmail.com} \\
}

\begin{document}

\maketitle

\begin{abstract}
Recent Large Language Models (LLMs) have significantly advanced natural language processing and automated decision-making. 
However, these models still encounter difficulties when performing complex reasoning tasks involving logical deduction and systematic planning, primarily due to their reliance on implicit statistical relationships without structured knowledge representation. 
Inspired by cognitive science and neurosymbolic AI, we introduce a novel approach to enhance LLMs through explicit structured reasoning.  
First, we convert unstructured data into structured formats by explicitly annotating reasoning steps. We then employ this structured dataset to train LLMs  through Supervised Fine-Tuning (SFT). Additionally, we enhance the structured reasoning capabilities of LLMs using Group Relative Policy Optimization (GRPO), incorporating two innovative algorithms—MAX-Flow and Longest Common Subsequence (LCS)—which notably improve reasoning effectiveness and reduce computational complexity. Experimental results from fine-tuning a DeepSeek-R1-Distill-Qwen-1.5B model demonstrate concise reasoning, robust performance across various scenarios, and improved compatibility with optimization techniques, validating the efficacy of structured reasoning integration in LLMs. Code is available at: \url{https://github.com/cnsdqd-dyb/Enhancing-Large-Language-Models-through-Structured-Reasoning}
\vspace{-10pt}
\end{abstract}
\section{Introduction}

Recent advancements in Large Language Models (LLMs), exemplified by models such as DeepSeek-R1~\cite{deepseek-r1}, OpenAI-o1~\cite{o1} and QwQ~\cite{noauthor_qwq-32b_nodate}, have impacted various domains including natural language processing, knowledge reasoning, and automated decision-making. 
Despite their remarkable abilities to generate fluent and contextually coherent text, LLMs usually demonstrate limitations when tackling complex reasoning tasks, such as logical deduction, systematic planning, and sophisticated problems in advanced mathematics and physics. These limitations stem primarily from their reliance on implicit, statistical associations learned from massive unstructured datasets, without explicit mechanisms to systematically represent and manipulate structured knowledge and logical relationships. 

Structured reasoning, a fundamental cognitive ability observed in human intelligence (\citep{bronkhorst2022students, forstmann2016sequential, evans2018dual, miller2001integrative}), involves explicitly representing problems, systematically manipulating knowledge, and drawing logical conclusions in a structured and interpretable manner. Integrating structured reasoning into LLMs has the potential to significantly improve their performance on tasks that require explicit logic and structured problem solving, bridging the gap between human-like reasoning and statistical text prediction. 

This paper proposes a novel approach to enhance LLMs with structured reasoning capabilities, inspired by cognitive science theories and recent advances in neurosymbolic artificial intelligence. Specifically, we introduce mechanisms that explicitly encode structured knowledge representations and reasoning processes in LLMs. Our method aims to leverage both the flexibility of neural networks and the interpretability and precision of symbolic reasoning.  

First, we transform unstructured data into structured data by incorporating explicit reasoning tags or labels that clearly indicate the current step of the reasoning process. 
In other words, we strengthen LLMs through structured reasoning with explicit step annotations. 
We utilize this structured dataset to train LLMs through Supervised Fine-Tuning (SFT).  
Additionally, we employ Group Relative Policy Optimization (GRPO) to further enhance structured reasoning. Specifically, GRPO integrates two key algorithms to promote effective and efficient structured reasoning: (1) \textit{MAX-Flow}, which surpasses traditional perplexity metrics in assessing the effectiveness of reasoning steps; and (2) \textit{Longest Common Subsequence (LCS)}, which preserves reasoning accuracy while substantially decreasing computational complexity. 
Using only 500 structured reasoning examples and 250 global reinforcement learning, we fine-tuned a DeepSeek-R1-Distill-Qwen-1.5B model. It achieves more concise reasoning, demonstrates stable performance in various scenarios, and offers enhanced compatibility with practical optimization methods.  
Our contributions are as follows,
\begin{itemize}
    \item We propose a novel Structured Reasoning approach, that achieves more concise  reasoning and stable performance across various scenarios, while enhancing the reasoning capabilities of Large Language Models (LLMs).

    \item We construct a structured reasoning dataset and demonstrate its effectiveness in enabling structured reasoning within LLMs through Supervised Fine-Tuning (SFT).

    \item We integrate MAX-Flow and Longest Common Subsequence (LCS) algorithms into Group Relative Policy Optimization (GRPO), reducing computational complexity while maintaining reasoning accuracy.

    \item Leveraging the interpretability advantages of structured reasoning, we present a novel approach to evaluate reasoning step metrics, while revealing the division of reasoning interests across neural layers, providing valuable insights for future LLM research.
\end{itemize}

\begin{figure}[t]
    \centering
    \includegraphics[width=\linewidth]{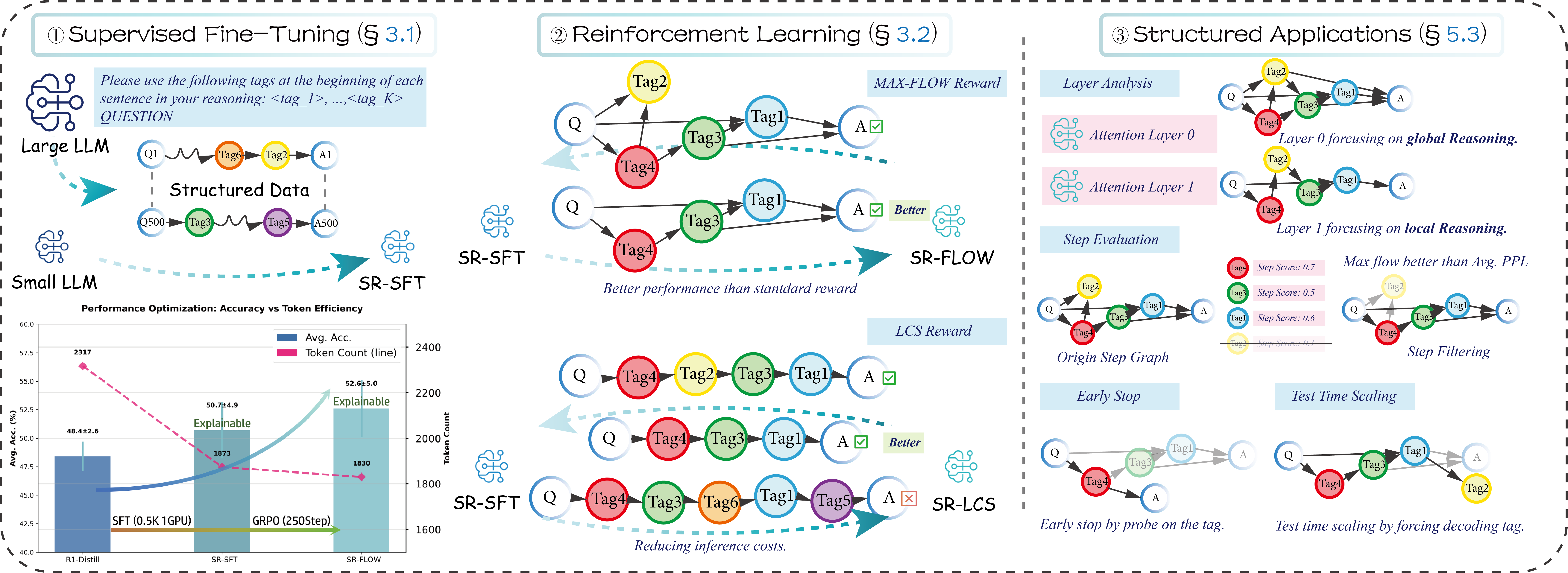}
    \caption{Illustration of enhancing Large Language Models (LLMs) with Structured Reasoning (SR) through (1) Supervised Fine-Tuning (SFT) and (2) Group Relative Policy Optimization (GRPO). In SFT, we construct a structured dataset that is  explicitly annotated with reasoning step labels. In GRPO, we apply the MAX-Flow algorithm to emphasize reasoning processes with balanced contributions, improving reasoning accuracy, and utilize the Longest Common Subsequence (LCS) algorithm to prioritize reasoning processes sharing common subsequences, thus reducing computational overhead. As demonstrated in (3), our approach results in more concise reasoning, stable performance across various settings, and improved compatibility with optimization methods.    
    }
    \label{fig:first_figure}
\end{figure}

\section{Related Work}
\textbf{Reinforcement Learning Helps Efficiency Improvement}
\label{sec:rl-based-efficiency}
Recent approaches use RL to improve reasoning efficiency, from basic length penalties~\citep{team2025kimi,li2025adaptive,arora2025training} to more sophisticated methods. L1~\citep{aggarwal2025l1} embeds length constraints in training instructions, while O1-Pruner~\citep{luo2025o1} balances brevity and accuracy against reference benchmarks. DAST~\citep{shen2025dast} introduces adaptive reasoning through token-length budget, allocating resources based on problem complexity. THINKPRUNE~\citep{hou2025thinkprune} employs a length-aware reward function with progressively tightening constraints, while Think When You Need~\citep{yang2025think} uses comparative rewards to guide models toward concise yet effective solutions.

\textbf{Efficient CoT According to Perplexity}
\label{sec:sft-based-efficiency}
Several works optimize reasoning chains using perplexity-based methods~\citep{perplexity}, including stepwise refinement~\citep{cui2025stepwise}, token pruning~\citep{xia2025tokenskip},  attack detection~\citep{alon2023detectinglanguagemodelattacks} and step elimination strategies~\citep{liu2024languagemodelslearnskip}. Furthermore, \citep{zhang2025entropybasedexplorationconductionmultistep} proposes exploration based on entropy for multistep reasoning. Our research reveals that perplexity metrics inadequately assess the importance of reasoning steps, demonstrating that our MAX-Flow algorithm outperforms perplexity-based approaches in evaluating the importance of reasoning steps.

\textbf{Language Model Reasoning (for Math)}
Since OpenAI-O1~\citep{jaech2024openai}, followed by O3~\citep{OpenAI2025o3mini} and DeepSeek-R1~\citep{deepseekai2025}, researchers have proposed increasingly sophisticated RL algorithms, including LCPO~\citep{aggarwal2025l1}, REINFORCE++~\citep{hu2025reinforce++}, DAPO~\citep{yu2025dapo}, DPO-VP~\citep{tu2025enhancing}, VinePPO~\citep{kazemnejad2024vineppo}, CPPO~\citep{lin2025cppo}, VAPO~\citep{vapo}, and GRO~\citep{cai2025one}. Empirical investigations have explored data scaling~\citep{shen2025exploring}, curriculum strategies~\citep{wen2025light,roux2025tapered}, and reward engineering~\citep{gao2024designing,cui2025process,ma2023eureka}. Recent evaluations~\citep{hochlehnert2025soberreasoning} show many reported improvements fail against properly optimized baselines. Our SR model incorporates structured reasoning through carefully designed SFT and reward approaches, across multiple seeds to ensure reproducibility.
\section{Method}


\subsection{Structured Supervised Fine-Tuning}
We transform unstructured data into structured data by incorporating explicit reasoning tags or labels that clearly indicate the current step of the reasoning process.  Initially, we establish a structured reasoning format and create a set of reasoning step annotations, followed by supervised fine-tuning.

\paragraph{Structured Reasoning Prompt}  
Given a question, we prompt the model as follows:
\begin{quote}
\texttt{Please use the following tags at the beginning of each sentence in your reasoning: <tag\_1>, ..., <tag\_K> QUESTION}.
\end{quote}
The model shows tagged reasoning, with unrestricted final answers:
\begin{quote}
\texttt{<think> <tag\_i> STEP1 (optional <\textbackslash tag\_i>) <tag\_j> STEP2 ... </think> ANSWER}.
\end{quote}
We first analyze the distribution of the tags used by DeepSeek-R1 671B(\citep{deepseek-r1}). After removing low-frequency tags, we retain 23 tags; the full prompts for tasks can be found in \ref{full_prompts}.

\paragraph{Supervised Fine-Tuning}  
We fine-tune the model to produce structured reasoning based on our designed prompt. For each Question-Reasoning-Answer triplet $(q, r, a)$ in the chosen dataset $\mathcal{D}_\mathrm{fine}$, the question $q$ is combined with our structured reasoning prompt $I$, which guides the model to employ designated reasoning tags at the beginning of each sentence. Subsequently, the model is trained to create the associated structured reasoning $r$ and provide the answer $a$. The fine-tuned model, $SR_\mathrm{sft}$, is trained as follows:
\begin{equation}
SR_\mathrm{sft} = \prod_{(q, r, a \in \mathcal{D}_\mathrm{fine}} P(r,a \mid q, I),
\end{equation}
where $I$ denotes our structured reasoning prompt and $\mathcal{D}_\mathrm{fine}$ is the set of selected high-quality samples.

\subsection{GRPO Rewards for Improved Structured Reasoning}

To further enhance the model's structured reasoning capabilities, we introduce two reasoning rewards to the GRPO (Group Relative Policy Optimization). We also truncated long outputs and shuffled prompt tags during training. The details can be found in \ref{experimental_setting}.

\paragraph{MAX-Flow Reward} 
To evaluate the importance of the reasoning step, we propose a reward metric based on the max-flow / min-cut theory (\citep{ford1956maximal}).

\textbf{1. Step-to-Step Attention Matrix.} 
\label{method:step-matrix}
\label{method:max-flow}
Given the attention tensor $\mathcal{A} \in \mathbb{R}^{H \times L_{seq} \times L_{seq}}$ from a certain layer, where $H$ denotes attention heads and $L_{seq}$ the sequence length, we compute the normalized step attention matrix $\mathbf{A} \in \mathbb{R}^{n \times n}$ for reasoning steps $n$. For step $i$ (token range $[s_i^\text{start}, s_i^\text{end}]$) step $j$:

\begin{equation}
A_{ij} = \frac{1}{H\mathcal{T}_i}\sum_{h=1}^{H}\sum_{a \in \mathcal{T}_i} \max_{b \in \mathcal{T}_j} \mathcal{A}_{h,a,b},
\end{equation}
where $\mathcal{T}_k \equiv s_k^\text{end} - s_k^\text{start}$ denotes the token length of step $k$.

\textbf{2. Graph-Based Flow Analysis.} 
Construct directed graph $G = (V,E)$.
Nodes $V = \{1,...,n\}$ representing steps (node $1$: Question, node $n$: Answer); Edges $(i,j) \in E$ with capacity $A_{ij}$ when $A_{ij} > \tau$ (threshold $\tau=0.05$).

\textbf{3. Importance Scoring.} 
For source $s=1$ and target $t=n$:
Compute max-flow $F$ in $G$ using Ford-Fulkerson algorithm.
For each node $k \in V \setminus\{s,t\}$, $\Delta F_k = F - F_{-k}$,
where $F_{-k}$ is the max-flow in subgraph $G_{-k}$ (node $k$ removed).
The value $\Delta F_k$ quantifies how crucial step $k$ is for reaching the conclusion.

\textbf{4. Robustness Evaluation.} 
The reasoning quality metric $Q \in [0,1]$ is computed as:
\begin{equation}
Q = 1 - \frac{\sum_{k \in \mathcal{K}_\text{top}} \Delta F_k}{\sum_{j=0}^{n-1} \Delta F_j},
\end{equation}
where $\mathcal{K}_\text{top}$ contains the top-25\% most important steps. Higher $Q$ indicates more balanced reasoning.

\paragraph{The Longest Common Subsequence Reward}
\label{method:lcs-reward}
Formally, given a set of reasoning completions \(\mathcal{R} = \{r_1, \ldots, r_n\}\) for a question, let \(r_{\mathrm{acc}}(r_i)\) denote the correctness reward for reasoning completion \(r_i\). For each pair \((r_i, r_j)\), we extract their reasoning steps and compute the longest common subsequence (LCS) of reasoning tags, denoted \(\mathrm{LCS}(r_i, r_j)\).

Let \(L_{\mathrm{lcs}}\) be the total length of the matched steps in the LCS and \(L_{i}\) be the total length of steps in \(r_i\).

To prevent \textit{length hacking} (i.e., artificially increasing the token count of each reasoning step to inflate scores), for each matched step \(k\) in the LCS with lengths \(\ell_{i,k}\) and \(\ell_{j,k}\), we introduce a length suppression factor, defined as \(\mathrm{ratio}_k = \frac{\ell_{j,k}}{2\ell_{i,k}}\) if \(\ell_{i,k} > \ell_{j,k}\) and \(\mathrm{ratio}_k = 1 - \frac{\ell_{i,k}}{2\ell_{j,k}}\) otherwise.
Subsequently, the length of the weighted LCS is defined as $L_{\mathrm{lcs}} = \sum_{k \in \mathrm{LCS}(r_i, r_j)} \mathrm{ratio}_k \cdot \ell_{i,k}.$

We define the pairwise LCS score as:
\begin{equation}
\mathrm{Score}_{\mathrm{lcs}}(r_i, r_j) =
\begin{cases}
\frac{L_{\mathrm{lcs}}}{L_{i}}, & \text{if both $r_i$ and $r_j$ are correct}, \\
- \frac{L_{\mathrm{lcs}}}{L_{i}}, & \text{if both $r_i$ and $r_j$ are incorrect}, \\
1 - \frac{L_{\mathrm{lcs}}}{L_{i}}, & \text{if $r_i$ is correct, $r_j$ is incorrect}, \\
-1 + \frac{L_{\mathrm{lcs}}}{L_{i}}, & \text{if $r_i$ is incorrect, $r_j$ is correct}.
\end{cases}
\end{equation}

Here, a higher LCS ratio is rewarded when compared with correct completions (encouraging consensus on high-quality reasoning), while a lower LCS ratio is rewarded when compared with incorrect completions (encouraging diversity from incorrect reasoning). The length suppression factor \(\mathrm{ratio}_k\) penalizes unnecessarily long steps and encourages concise reasoning.

Finally, the overall LCS reasoning reward for \(c_i\) is averaged over all other completions:
\begin{equation}
r_{\mathrm{lcs}}(c_i) = \frac{1}{n-1} \sum_{j \neq i} \mathrm{Score}_{\mathrm{lcs}}(c_i, c_j).
\end{equation}

\section{Experiments}

\textbf{Datasets.}
In the first stage, we use the S1 dataset~\citep{muennighoff2025s1simpletesttimescaling}, which contains 1,000 high-quality and diverse problems with reference answers and reasoning traces, covering science, technology, engineering and mathematics (STEM) and related domains. For both structured and unstructured reasoning fine-tuning data, we select the 500 most challenging problems (with the lowest accuracy) and generate correct samples from DeepSeek R1.
In the second stage, we fine-tune on the DeepScaleR-Preview-Dataset~\citep{deepscaler2025}, a mathematics dataset containing 40K question-answer pairs drawn from AIME, AMC, Omni-Math~\citep{gao2024omnimathuniversalolympiadlevel}, and STILL~\citep{R1-searcher}. 

\textbf{Evaluation.} We evaluate our models on the test sets of nine reasoning datasets: AIME 2024, AIME 2025, AMC, MATH500~\citep{hendrycksmath2021}, Minerva, Olympiad-Bench~\citep{he2024olympiadbench}, and out-of-domain benchmarks including  GPQA-Diamond~\citep{rein2023gpqagraduatelevelgoogleproofqa}, LSAT-AR~\citep{zhong2023agievalhumancentricbenchmarkevaluating},  MMLU-ALL-VALID~\citep{hendrycks2021measuringmassivemultitasklanguage}.
Following the recommendations of Sober Reasoning~\citep{hochlehnert2025soberreasoning}, we use three different random seeds for sampling and evaluation on large datasets MATH500, Minerva, and Olympiad-Bench, reporting both the mean and variance. For small datasets, we use ten different seeds to ensure statistical reliability.

\textbf{Models.} Our base model is DeepSeek-R1-Distill-Qwen-1.5B~\citep{deepseekai2025deepseekr1incentivizingreasoningcapability}. Due to computational constraints, we restrict the maximum context length to 4K tokens during supervised fine-tuning (SFT) using the 500 samples from the S1 dataset, obtaining the \textbf{SR-SFT} model.

We then derive three variants through reinforcement learning fine-tuning:
\begin{itemize}
\item \textbf{SR-ACC}: trained for 250 steps using the standard accuracy reward.
\item \textbf{SR-LCS}: trained for 250 steps using the LCS reward (Equation~\ref{method:lcs-reward}).
\item \textbf{SR-FLOW}: trained for 250 steps using our MAX-Flow reward (Equation~\ref{method:max-flow}).
\end{itemize}
The structured reasoning example can be found in \ref{reasoning example}. The experimental setting can be found in \ref{experimental_setting}.

\textbf{Baselines.} We compare our proposed methods with the following baselines:
Qwen2.5-Math-1.5B, Qwen2.5-Math-1.5B-Instruct~\citep{yang2024qwen25mathtechnicalreportmathematical}
Qwen2.5-Math-1.5B-Oat-Zero~\citep{liu2025oatzero}
These models are all initialized from Qwen2.5 Math 1.5B and subsequently finetuned via reinforcement learning.
DeepSeek-R1-Distill 1.5B~\citep{deepseekai2025deepseekr1incentivizingreasoningcapability}, 
FastCurl-1.5B-Preview~\citep{song2025fastcurlcurriculumreinforcementlearning},
DeepScaleR-1.5B~\citep{deepscaler2025},
II-1.5B-Preview,
STILL-3-1.5B~\citep{min2024imitateexploreselfimprovereproduction},
OpenRS1-1.5B, OpenRS2-1.5B, and OpenRS3-1.5B~\citep{dang2025reinforcementlearningreasoningsmall},
L1-Qwen-1.5B-Max, L1-Qwen-1.5B-Exact~\citep{aggarwal2025l1controllinglongreasoning}). Note that these models and our model are all initialized from DeepSeek-R1-Distill-1.5B and subsequently fine-tuned via reinforcement learning (e.g. GRPO~\citep{shao2024deepseekmathpushinglimitsmathematical}) to enhance mathematical reasoning capabilities.

\section{Results}

\subsection{Main Experiment Results.}
In this subsection, we evaluate the effectiveness of our proposed methods by reporting Pass@1 accuracy (mean ± standard deviation) across six math benchmarks using standardized evaluation protocols. For AIME24, AIME25 and AMC23, we perform evaluations in 10 seeds each, while MATH500, Minerva and OlympiadBench are evaluated in 3 seeds each. We report two aggregate metrics: the average score across all six benchmarks (Avg. score) and more stable large average score (Large Avg.) computed from the three benchmarks with lower variance (MATH500, Minerva, and OlympiadBench). We perform comprehensive comparisons between the base model, our SR model, and the standard SFT model across both in-domain and out-of-domain datasets (MATH500, Minerva, OlympiadBench, GPQA-Diamond, LSAT-AR, MMLU-ALL-VALID) using 8K context length. 
Additionally, for correct completions of MATH500, we analyze the average token length and the average number of reasoning steps to characterize the reasoning process.

\vspace{-10pt}
\begin{table}[htbp]
\centering
\caption{Training Details. To ensure consistency in counting training steps, we standardized the batch size to 128. This means that two steps with a batch size of 64 are considered equivalent to one step with a batch size of 128. For SR, we performed 5 epochs of SFT followed by 250 steps of GRPO.}
\resizebox{\textwidth}{!}{
\begin{tabular}{lccc}
\hline
\textbf{Model} & \textbf{Training Steps} & \textbf{Training Stages} & \textbf{Number of GPUs Used in Each Stage} \\
\hline
SR & SFT+RL($\sim$ 23) & 2 & 1, 4 \\
FastCuRL & RL($\sim$ 860) & 4 & 8, 8, 8, 8 \\
DeepScaleR & RL($\sim$ 1,750) & 3 & 8, 16, 32 \\
\hline
\end{tabular}
}
\label{tab:training_details}
\end{table}
\vspace{-10pt}

\paragraph{Competitive Performance at Minimal Cost.}
After fine-tuning, our model achieves a high performance as shown in Table~\ref{tab:benchmark_results}, improving the average score from 48.4 to 50.7 and the large benchmark average from 55.9 to 56.7. Building upon our MAX Flow Reward method (Section~\ref{method:max-flow}), we further trained the model using GRPO, achieving additional performance gains, increasing the average score from 50.7 to 52.6 and the large benchmark average from 56.7 to 58.1. Our experiments validate that structured reasoning LLMs can rapidly approach the performance of current state-of-the-art models (trained with multistage 8k-64k long-context large-scale datasets; see Table~\ref{tab:training_details}) when trained efficiently with: (1) limited amounts of high-quality structured data and (2) reward functions specifically designed for structured reasoning.

\begin{table}[htbp]
\centering
\caption{Benchmark Results (Pass@1 Accuracy). All results are reported as mean ± standard deviation. Avg. score calculates the average across all six benchmarks, while Large Avg. focuses on the more stable MATH500, Minerva, and Olympiad benchmarks. Top-3 models in each category are highlighted with increasing gray intensity.}
\resizebox{\textwidth}{!}{
\begin{tabular}{lcccccccc}
\hline
\textbf{Model} & \textbf{AIME'24} & \textbf{AIME'25} & \textbf{AMC'23} & \textbf{MATH500} & \textbf{Minerva} & \textbf{Olympiad} & \textbf{Avg.} & \textbf{Large Avg.} \\
\hline
\multicolumn{9}{l}{\textbf{Based on: Qwen2.5-Math-1.5B (RL)}} \\
Math & 11.3±\footnotesize{3.6} & 5.7±\footnotesize{2.7} & 44.0±\footnotesize{4.9} & 51.7±\footnotesize{5.5} & 11.3±\footnotesize{2.2} & 26.0±\footnotesize{0.6} & 25.0±\footnotesize{3.3} & 29.7±\footnotesize{2.8} \\
Oat-Zero & 16.0±\footnotesize{3.2} & 6.7±\footnotesize{3.4} & 52.5±\footnotesize{2.9} & 73.5±\footnotesize{1.7} & 26.3±\footnotesize{0.8} & 37.2±\footnotesize{1.3} & 32.0±\footnotesize{2.2} & 45.7±\footnotesize{1.3} \\
Math & 12.0±\footnotesize{1.7} & 11.7±\footnotesize{5.7} & 54.8±\footnotesize{5.3} & 74.7±\footnotesize{0.5} & 26.7±\footnotesize{1.8} & 37.9±\footnotesize{0.2} & 36.3±\footnotesize{2.5} & 46.4±\footnotesize{0.8} \\
\hline
\multicolumn{9}{l}{\textbf{Based on: Deepseek-R1-Distill-Qwen-1.5B (RL)}} \\
R1-Distill & 28.7±\footnotesize{4.8} & 22.3±\footnotesize{5.2} & 71.5±\footnotesize{3.9} & 84.9±\footnotesize{0.3} & 30.5±\footnotesize{1.0} & 52.4±\footnotesize{0.4} & 48.4±\footnotesize{2.6} & 55.9±\footnotesize{0.6} \\
L1-Exact & 24.4±\footnotesize{3.3} & 22.3±\footnotesize{4.2} & 70.5±\footnotesize{3.7} & \cellcolor{gray!20}86.6±\footnotesize{0.8} & 31.5±\footnotesize{1.7} & 52.5±\footnotesize{1.3} & 47.9±\footnotesize{2.5} & 56.9±\footnotesize{1.3} \\
L1-Max & 27.7±\footnotesize{4.2} & 21.0±\footnotesize{5.0} & 73.2±\footnotesize{6.0} & 84.7±\footnotesize{0.1} & \cellcolor{gray!30}33.3±\footnotesize{0.9} & 52.3±\footnotesize{0.6} & 48.7±\footnotesize{2.8} & 56.8±\footnotesize{0.5} \\
Open-RS1 & 28.9±\footnotesize{6.0} & 21.3±\footnotesize{4.2} & 75.0±\footnotesize{3.3} & 85.1±\footnotesize{0.8} & 30.4±\footnotesize{0.2} & 53.2±\footnotesize{1.9} & 49.0±\footnotesize{2.7} & 56.2±\footnotesize{1.0} \\
Open-RS2 & 31.3±\footnotesize{7.7} & 22.7±\footnotesize{5.6} & 73.0±\footnotesize{5.7} & 84.1±\footnotesize{0.2} & 29.2±\footnotesize{1.1} & 53.7±\footnotesize{0.6} & 49.0±\footnotesize{3.5} & 55.7±\footnotesize{0.6} \\
Open-RS3 & 29.7±\footnotesize{4.6} & 24.7±\footnotesize{6.5} & 69.2±\footnotesize{5.5} & 84.2±\footnotesize{1.1} & 28.6±\footnotesize{2.3} & 51.8±\footnotesize{0.8} & 48.0±\footnotesize{3.5} & 54.9±\footnotesize{1.4} \\
STILL-3 & 34.7±\footnotesize{5.5} & 24.0±\footnotesize{6.4} & 72.5±\footnotesize{5.4} & \cellcolor{gray!20}86.6±\footnotesize{1.9} & 30.0±\footnotesize{0.6} & 53.9±\footnotesize{1.5} & 50.3±\footnotesize{3.6} & 56.8±\footnotesize{1.3} \\
II-Thought & 32.0±\footnotesize{5.9} & 24.0±\footnotesize{4.1} & \cellcolor{gray!60}79.5±\footnotesize{5.1} & \cellcolor{gray!20}86.6±\footnotesize{0.6} & \cellcolor{gray!20}31.7±\footnotesize{0.6} & \cellcolor{gray!30}54.9±\footnotesize{0.4} & 51.5±\footnotesize{2.8} & \cellcolor{gray!20}57.7±\footnotesize{0.5} \\
FastCuRL & \cellcolor{gray!20}36.3±\footnotesize{4.3} & \cellcolor{gray!30}27.0±\footnotesize{3.7} & \cellcolor{gray!30}78.8±\footnotesize{4.1} & \cellcolor{gray!60}87.9±\footnotesize{1.2} & 30.8±\footnotesize{1.4} & \cellcolor{gray!60}56.5±\footnotesize{0.6} & \cellcolor{gray!20}52.9±\footnotesize{2.6} & \cellcolor{gray!60}58.4±\footnotesize{1.1} \\
DeepScaleR & \cellcolor{gray!60}37.0±\footnotesize{6.6} & \cellcolor{gray!60}30.3±\footnotesize{4.3} & 76.2±\footnotesize{4.6} & \cellcolor{gray!30}87.8±\footnotesize{1.0} & 31.0±\footnotesize{1.5} & 55.5±\footnotesize{1.1} & \cellcolor{gray!60}53.0±\footnotesize{3.2} & \cellcolor{gray!30}58.1±\footnotesize{1.2} \\
\hline
\multicolumn{9}{l}{\textbf{Ours Based on: Deepseek-R1-Distill-Qwen-1.5B (RL)}} \\
SR-SFT & 32.7±\footnotesize{8.7} & \cellcolor{gray!20}25.3±\footnotesize{8.0} & 75.8±\footnotesize{6.7} & 85.6±\footnotesize{1.6} & 31.3±\footnotesize{2.8} & 53.3±\footnotesize{1.9} & 50.7±\footnotesize{4.9} & 56.7±\footnotesize{2.1} \\
SR-FLOW & \cellcolor{gray!30}36.7±\footnotesize{8.9} & \cellcolor{gray!30}27.0±\footnotesize{8.2} & \cellcolor{gray!20}77.8±\footnotesize{6.6} & 85.3±\footnotesize{1.6} & \cellcolor{gray!60}34.2±\footnotesize{2.9} & \cellcolor{gray!20}54.9±\footnotesize{1.9} & \cellcolor{gray!20}52.6±\footnotesize{5.0} & \cellcolor{gray!30}58.1±\footnotesize{2.1} \\
\hline
\end{tabular}
}
\label{tab:benchmark_results}
\end{table}

\begin{figure}[ht]
    \centering
    \includegraphics[width=1.0\textwidth]{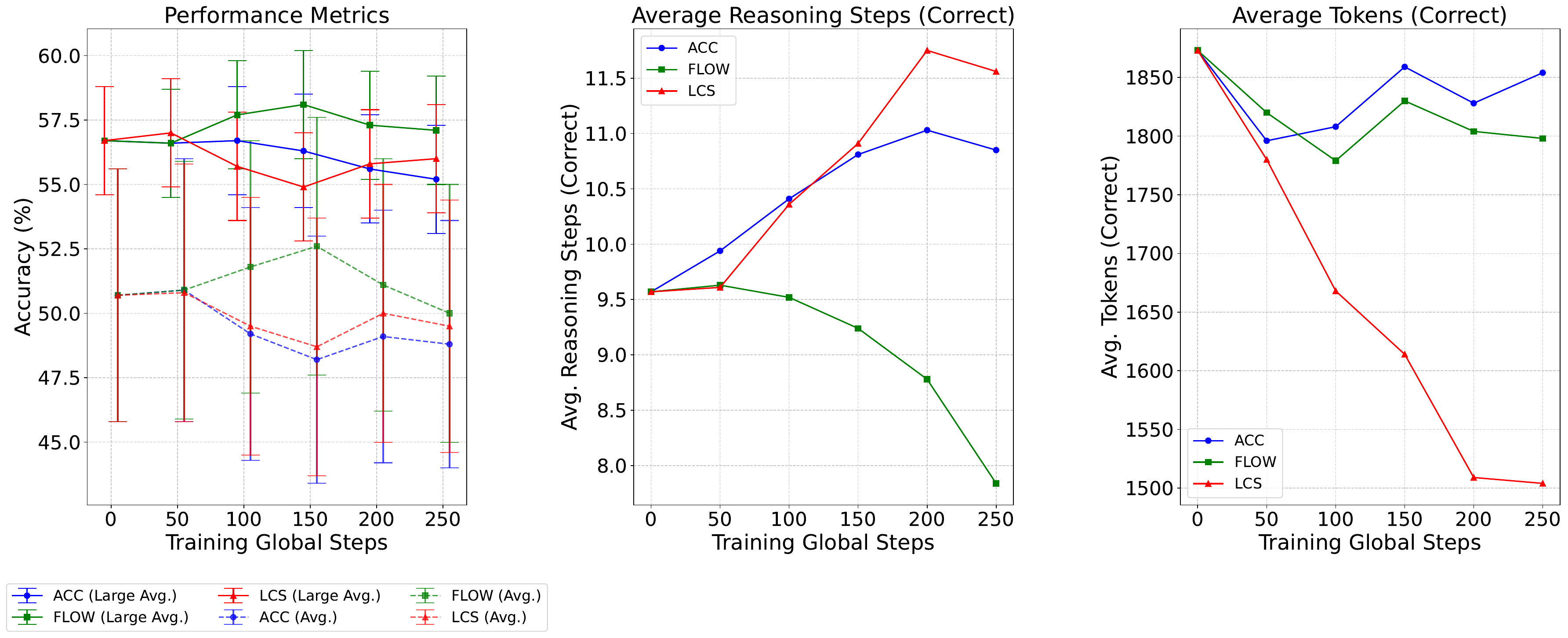}
    \caption{Comparison across Global Steps: Pass@1 Performance (Left), Reasoning Steps (Middle), and Average Tokens (Right) for \textcolor{blue}{SR-ACC}(standard reward), \textcolor{teal}{SR-FLOW}, and \textcolor{red}{SR-LCS}.}
    \label{fig:training_curve}
\end{figure}

\vspace{-20pt}
\paragraph{Reward Analysis During Training.}
Figure~\ref{fig:training_curve} and Table~\ref{tab:methods_comparison} presents the learning dynamics of our models during RL fine-tuning, with evaluations conducted at 50-step intervals. The curves demonstrate the evolution of: (1) SR model performance (both Avg. and Large Avg. scores), (2) the average number of reasoning steps in correct completions on MATH500, and (3) the average token length of solutions on MATH500.

\vspace{-10pt}
\textbf{Reasoning Efficiency Improvements.} Our proposed MAX-Flow reward and LCS reward offer different advantages in the trade-off between higher performance and better reasoning efficiency.

As shown in Figure~\ref{fig:training_curve} and Table~\ref{tab:methods_comparison}, SR-FLOW achieves \textbf{Higher Accuracy:} +1.9\% average gain on the "Larger Avg." metric at 150 steps compared to SR-ACC. \textbf{More Concise Reasoning:} Steps reduced from 9.57 to 7.84 while maintaining 84.7\% accuracy on MATH500. We observed that although we did not explicitly require the model to reduce the reasoning steps or token count, SR-FLOW spontaneously began to perform the fusion of the reasoning steps.

SR-LCS achieves \textbf{Token Efficiency:} 1504 tokens vs. 1873 (baseline) with a slight drop in Large Avg. from 56.7 to 56.0. Figure~\ref{fig:lcs_dist} shows more responses in the 256-1024 token range and fewer that exceed 8192 tokens, indicating more efficient reasoning. With accuracy rates of 84.4\%, 84.3\%, and 84.5\% for LCS-50, LCS-150, and LCS-250 respectively, the model maintains performance while improving efficiency under 8k context length constraints.

\begin{table}[ht]
\centering
\caption{Comprehensive Comparison on In-domain and Out-of-domain Data with 8k Context Length.}
\label{tab:model_comparison}
\resizebox{\textwidth}{!}{
\begin{tabular}{lcccccc}
\toprule
\textbf{Method} & \textbf{MATH500} & \textbf{Minerva} & \textbf{OlympiaBench} & \textbf{GPQA} & \textbf{LSAT-AR} & \textbf{MMLU} \\
\midrule
R1-Distill-1.5B (Base) & 80.33\%\footnotesize{±1.78\%} & 31.00\%\footnotesize{±2.81\%} & 44.49\%\footnotesize{±1.91\%} & 35.02\%\footnotesize{±3.40\%} & 26.26\%\footnotesize{±2.90\%} & 50.60\%\footnotesize{±1.28\%} \\
SFT (Standard) & 82.00\%\footnotesize{±1.72\%} & 28.68\%\footnotesize{±2.75\%} & 48.84\%\footnotesize{±1.93\%} & 33.66\%\footnotesize{±3.34\%} & 25.87\%\footnotesize{±2.89\%} & 44.83\%\footnotesize{±1.27\%} \\
\hline

SR-SFT & 84.53\%\footnotesize{±1.62\%} & 32.11\%\footnotesize{±2.84\%} & 49.53\%\footnotesize{±1.93\%} & 36.20\%\footnotesize{±3.39\%} & 26.00\%\footnotesize{±2.89\%} & 46.66\%\footnotesize{±1.28\%} \\
SR-Flow (150 steps) & 84.74\%\footnotesize{±1.58\%}\footnotesize{\textcolor{teal}{+0.25\%}} & 33.52\%\footnotesize{±2.88\%}\footnotesize{\textcolor{teal}{+4.39\%}} & 50.37\%\footnotesize{±1.92\%}\footnotesize{\textcolor{teal}{+1.70\%}} & 36.70\%\footnotesize{±3.43\%}\footnotesize{\textcolor{teal}{+1.38\%}} & 26.78\%\footnotesize{±2.92\%}\footnotesize{\textcolor{teal}{+3.00\%}} & 46.81\%\footnotesize{±1.28\%}\footnotesize{\textcolor{teal}{+0.32\%}} \\
\bottomrule
\end{tabular}
}
\end{table}

We also conducted a comprehensive evaluation on both in-domain and out-of-domain benchmarks (Table~\ref{tab:model_comparison}). During the reinforcement learning stage, our SR-Flow outperforms SR-SFT across all benchmarks (ranging from $+0.25\%$ to $+4.39\%$), validating the universal advantage of our approach in enhancing structured reasoning capabilities. As expected, the SFT model shows improvements on in-domain tasks but degradation on out-of-domain benchmarks due to domain shifting.

\subsection{What Are the Advantages of Structured Reasoning?}


\paragraph{Structured Reasoning Naturally Produces More Concise Token Length Distributions.} 

For large models, Table~\ref{tab:large_model_lengths} summarizes the average token lengths and accuracies across several benchmarks, including MATH500, GPQA-Diamond, MMLU-ALL-VALID, AMC23, and AIME24. Notably, the structured reasoning model achieves similar or higher accuracy with much shorter answers, e.g., on MATH500, the average reasoning token length drops from 2945 (base) to 1577 (SR-Prompt), while accuracy remains above 92\% (See Appendix \ref{reasoning prompt}  for the fill in the middle prompting strategy).

\vspace{-10pt}
\begin{table}[ht]
\centering
\caption{Token length and accuracy comparison for DeepSeek-R1 671B on several benchmarks.}
\label{tab:large_model_lengths}
\resizebox{\textwidth}{!}{
\begin{tabular}{lcccc}
\toprule
\textbf{Model} & \textbf{MATH500 Acc./Len.} & \textbf{GPQA Acc./Len.} & \textbf{MMLU Acc./Len.} & \textbf{AMC23 Acc./Len.} \\
\midrule
Base & 92.9\%/2945 & 70.3\%/6537 & 88.8\%/989 & 100\%/1716 \\
SR-Prompt & 93.0\%/1577\footnotesize{\textcolor{teal}{-46.5\%}} & 71.1\%/4028\footnotesize{\textcolor{teal}{-38.4\%}} & 89.6\%/512\footnotesize{\textcolor{teal}{-48.2\%}} & 100\%/2053\footnotesize{\textcolor{red}{+19.6\%}} \\
\bottomrule
\end{tabular}
}
\end{table}
\vspace{-5pt}

\begin{figure}[htbp]
    \centering
    \begin{subfigure}[b]{0.495\textwidth}
        \centering
        \includegraphics[width=\linewidth]{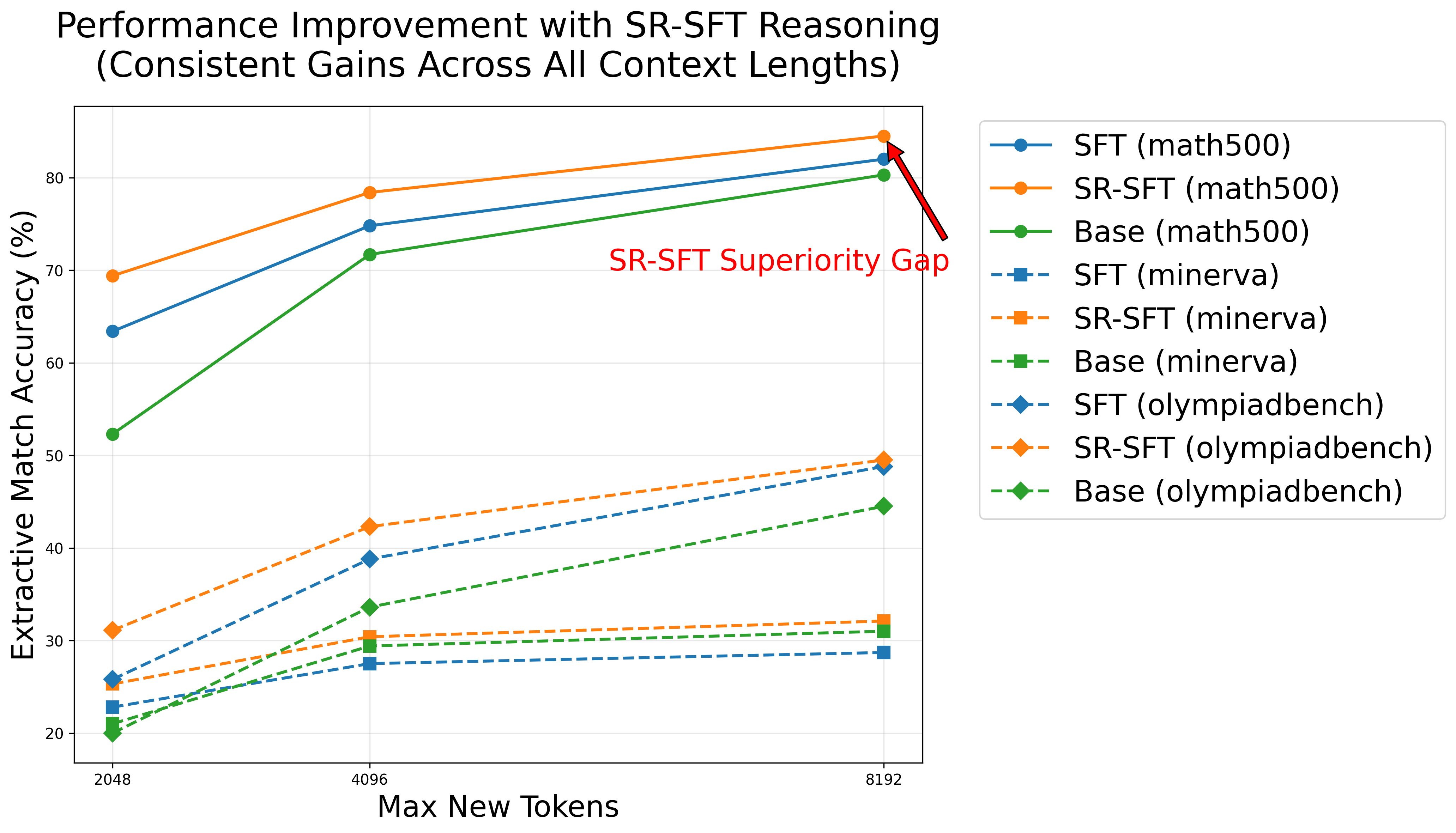}
        \caption{Accuracy across 2k, 4k, 8k Max New Tokens.}
        \label{fig:model_compare}
    \end{subfigure}
    \hfill 
    \begin{subfigure}[b]{0.495\textwidth}
        \centering
        \includegraphics[width=\linewidth]{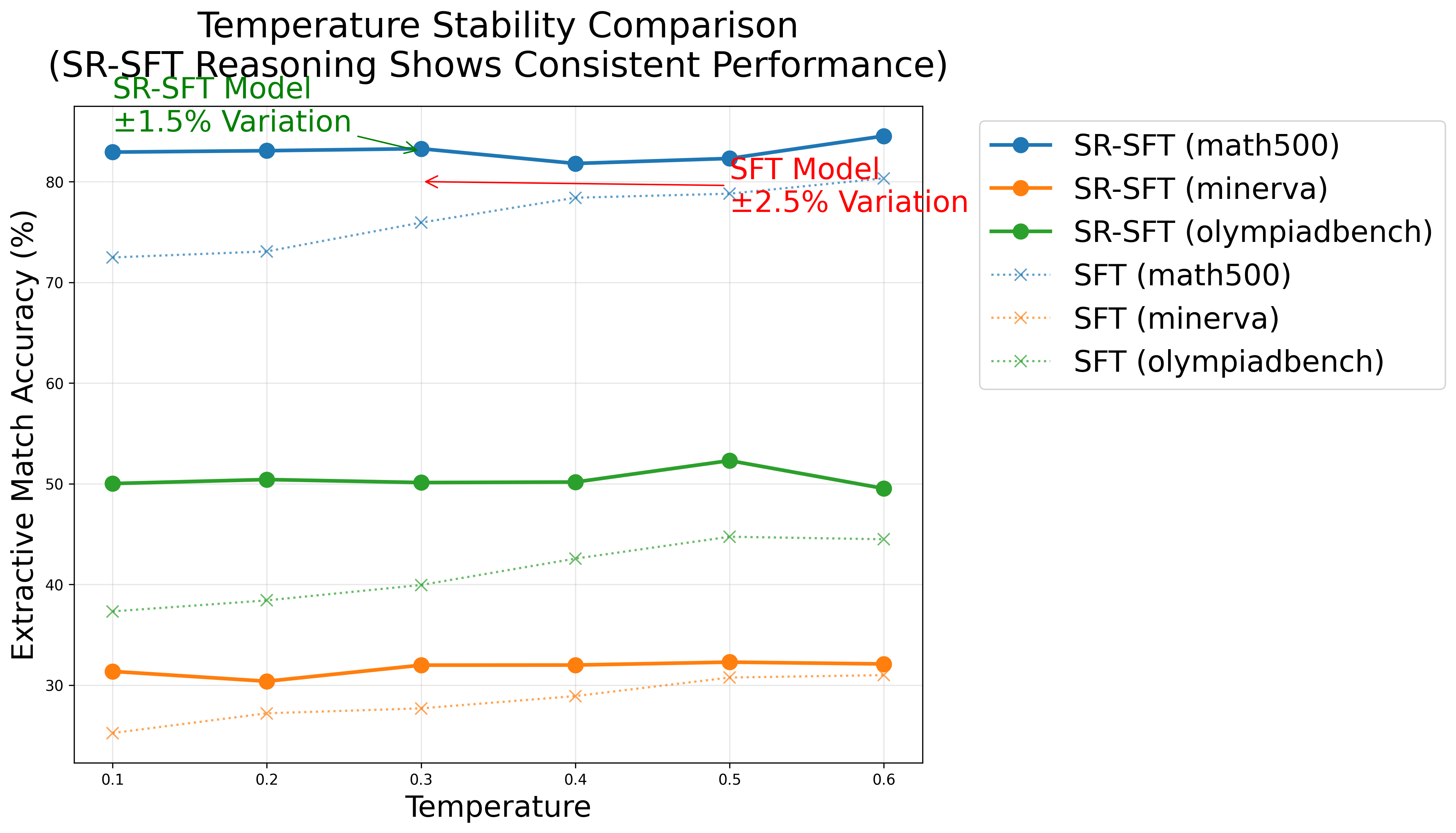}
        \caption{Temperature robustness comparison.}
        \label{fig:temp_stability}
    \end{subfigure}
    
    \caption{Performance Comparison of 1.5B Models across Maximum New Tokens (Left) and Temperature Settings (Right), averaged over three seeds.}

    \label{fig:combined_results}
\end{figure}

For small models, we constructed structured and unstructured reasoning datasets using the same problems from s1k for fine-tuning and analyzed the model outputs truncated at 2k, 4k, and 8k tokens. Figure~\ref{fig:model_compare} shows that models fine-tuned on structured data (SR-SFT) have better performance across all token limits (2k, 4k, and 8k) compared to standard SFT. Specifically, at the 8k limit, the unstructured model had 17.0\% of responses truncated with an average correct reasoning length of 2058 tokens, while the standard SFT model had only 10\% truncated with an average length of 1735 tokens, demonstrating a natural tendency to produce more concise answers.

\vspace{-5pt}

\paragraph{Structured Reasoning Models Produce More Stable Outputs.}
As shown in Figure~\ref{fig:temp_stability}, we observed an interesting phenomenon across experiments conducted with 3 seeds per task using an 8k context length. When the sampling temperature increases from 0 to 0.6 (0.6 is the recommended temperature from DeepSeek), the unstructured SFT model's accuracy rises significantly on all datasets: MATH500 (from 72.47 to 80.33), Minerva (from 25.25 to 31.00), and OlympiadBench (from 37.33 to 44.49). In contrast, our SR-SFT model maintains more consistent accuracy across temperature values: MATH500 (from 82.93 to 84.53), Minerva (from 31.37 to 32.11), and OlympiadBench (from 50.02 to 49.53).
This temperature sensitivity is much more pronounced in unstructured reasoning models, while structured reasoning models demonstrate stability across the temperature range.

\vspace{-5pt}
\begin{table}[htbp]
\centering
\caption{Comparing Trigger Counts and Distances to First Correct Answer Across Different Methods.}
\resizebox{\columnwidth}{!}{
\begin{tabular}{llcc}
\hline
\textbf{Trigger Type} & \textbf{Settings} & \textbf{Avg. Trigger Count} ↓ & \textbf{Avg. Distance to First Correct Answer (tokens)} ↓ \\
\hline
\textbf{Top tags} & "verify", "summarize", etc. & \textbf{2.02} & \textbf{78.01} \\
Token chunks & 128-token intervals & 3.93 & 131.05 \\
Keywords & "but", "wait", "however", etc. & 2.69 & 139.97 \\
\hline
\end{tabular}
}
\label{tab:missed_detection}
\end{table}

\paragraph{Improved Compatibility with Test-time Scaling and Early Stopping.}
For Test-time Scaling, existing work extends model outputs by injecting prompt tokens at thought-stopping points. Our method simplifies this by guiding outputs through the most likely next tag at stopping points. 
For early stopping, our tag-based approach outperforms traditional methods. In our experiment with 705 correct MATH500 reasoning completions (Table~\ref{tab:trigger_stats}), we compared interval-based (128-token), keyword-based ("but", "wait", "however", etc.), and tag-based ("verify", "summarize", etc.) detection strategies. 
As Table~\ref{tab:missed_detection} shows, our structured approach reduces average Probe-In-Middle interventions to just 2.02 while maintaining closest proximity to correct answers (78.01 tokens).


    

\vspace{-5pt}
\subsection{What Are the Gains from Structured Reasoning Models?}
\paragraph{Structured Analysis Helps Identify Redundant Reasoning Steps.}
\label{exp:IISR}
We designed an experiment (Interference Injection and Selective Removal, \textbf{IISR})~(\ref{ablation:IISR}) to assess whether we can better analyze the importance of reasoning steps. Since existing datasets rarely provide direct importance annotations for reasoning steps, and LLM-based scoring is noisy, we injected obviously irrelevant reasoning steps into existing chains. While we cannot confirm the relative importance of original steps, we are certain about the irrelevance of the injected ones. We compared our max-flow algorithm~(\ref{method:max-flow}), top-p/top-k backtracking~(\ref{other algorithms}), average step perplexity~(\ref{other algorithms}), and random selection for their Error Filtering Efficiency~(\ref{efe formula}) when removing 1-11 steps from mixed reasoning chains.

This experiment used 70 correctly reasoned examples from S1k covering science, technology, engineering, and mathematics (STEM) and related domains with longer trace lengths and more uniform reasoning steps. We defined four types of interference prompts: (1) \textbf{Redundant} - statements like ``Let's summarize what we've done so far, our previous work is correct'' that add no value to reasoning; (2) \textbf{Distracted} - comments indicating distraction such as ``This reminds me of another problem''; (3) \textbf{Harmful} - randomly injected reasoning steps from other problems; and (4) \textbf{Confused} - copies of current reasoning steps randomly injected at incorrect positions in the reasoning chain.

Through extensive randomized experiments, we found that as more reasoning steps were removed, our proposed methods based on step-matrix (See Section~\ref{method:step-matrix}) (top-k, top-p, and max-flow) significantly outperformed random removal. The specific example can be found in Appendix~\ref{ablation:IISR}.

\begin{tcolorbox}[
  enhanced,
  title={\bfseries Takeaway 1},
  colback=white,
  colframe=yellow!70!black,
  colbacktitle=yellow!70!black,
  coltitle=white,
  boxrule=1pt,
  arc=4mm,
  fonttitle=\bfseries
]
Using average perplexity of reasoning steps as a reasoning step evaluation metric is \textbf{unreliable}.
\end{tcolorbox}

Additionally, in our comparison with perplexity-based algorithms, we found that removing steps with the lowest PPL (PPL Bottom) performed similarly (though slightly worse) to our methods when dealing with redundant but harmless information, as such information typically has low information content and low perplexity. Interestingly, for logically confused interference, removing steps with the highest PPL (PPL Top) performed slightly better, as steps appearing in inappropriate positions caused significantly increased perplexity. This shows that PPL primarily reflects information quantity and cannot distinguish valuable reasoning from disruptive content. Our step-matrix-based methods significantly outperformed PPL-based approaches across all tasks.

\paragraph{Reasoning Models may Exhibit Distinct Layer-wise Division of Reasoning Roles.}
Based on our Top-P reasoning step filtering algorithm, which showed excellent performance in the IISR experiment, we further compared the changes in Error Filtering Efficiency across 0-28 layers.
\begin{tcolorbox}[
  enhanced,
  title={\bfseries Takeaway 2},
  colback=white,
  colframe=yellow!70!black,
  colbacktitle=yellow!70!black,
  coltitle=white,
  boxrule=1pt,
  arc=4mm,
  fonttitle=\bfseries
]
During the reasoning process, the earlier layers of small models \textbf{alternate} between focusing on local and global reasoning, while later layers predominantly attend to global reasoning.
\end{tcolorbox}

\begin{figure}[ht]
    \centering
    \begin{subfigure}[b]{0.49\textwidth}
        \includegraphics[width=\textwidth]{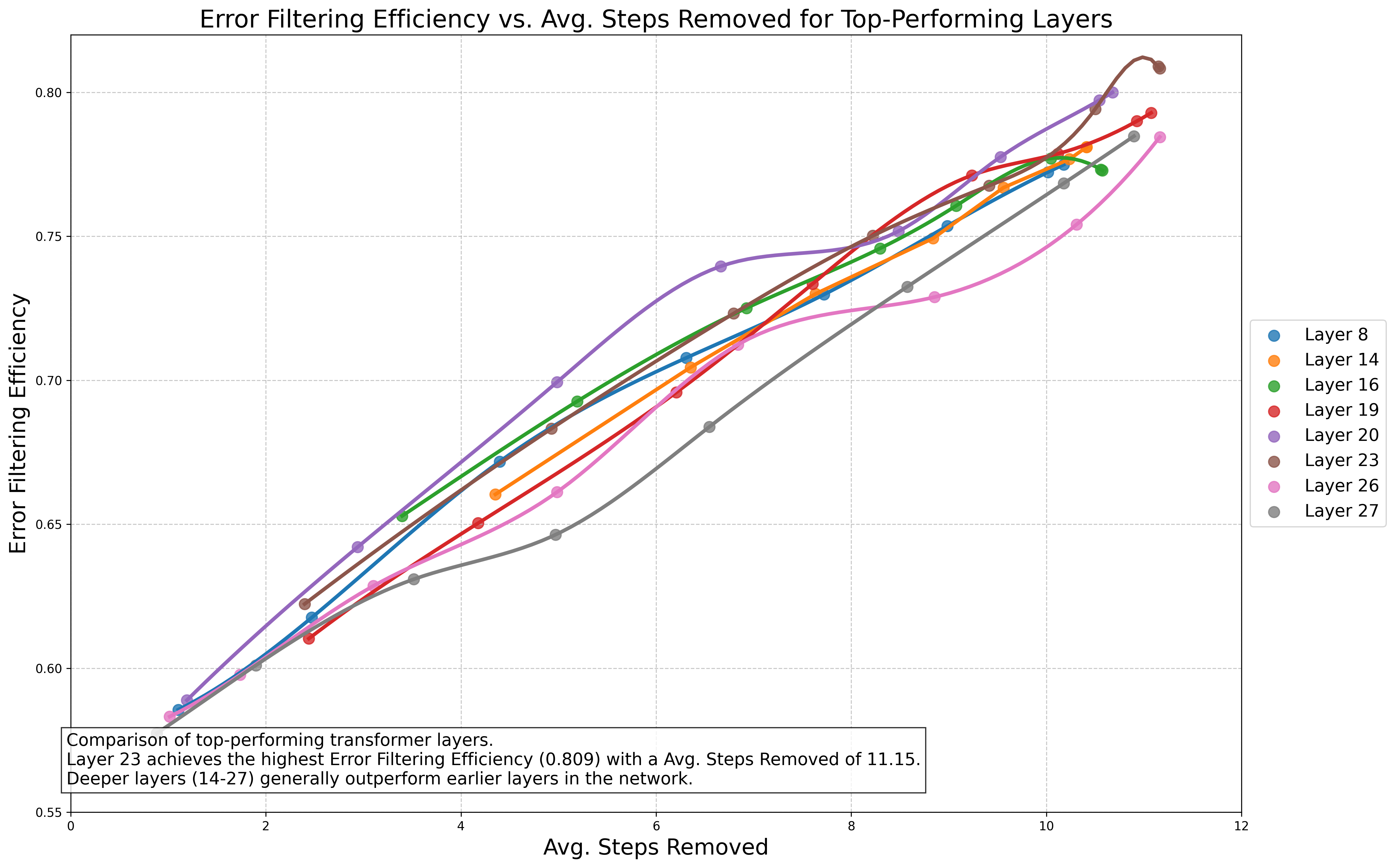}
        \caption{Comparison of Error Filtering Efficiency across different model layers, demonstrating how later layers better capture reasoning step importance.}
        \label{fig:best_layer}
    \end{subfigure}
    \hfill
    \begin{subfigure}[b]{0.50\textwidth}
        \includegraphics[width=\textwidth]{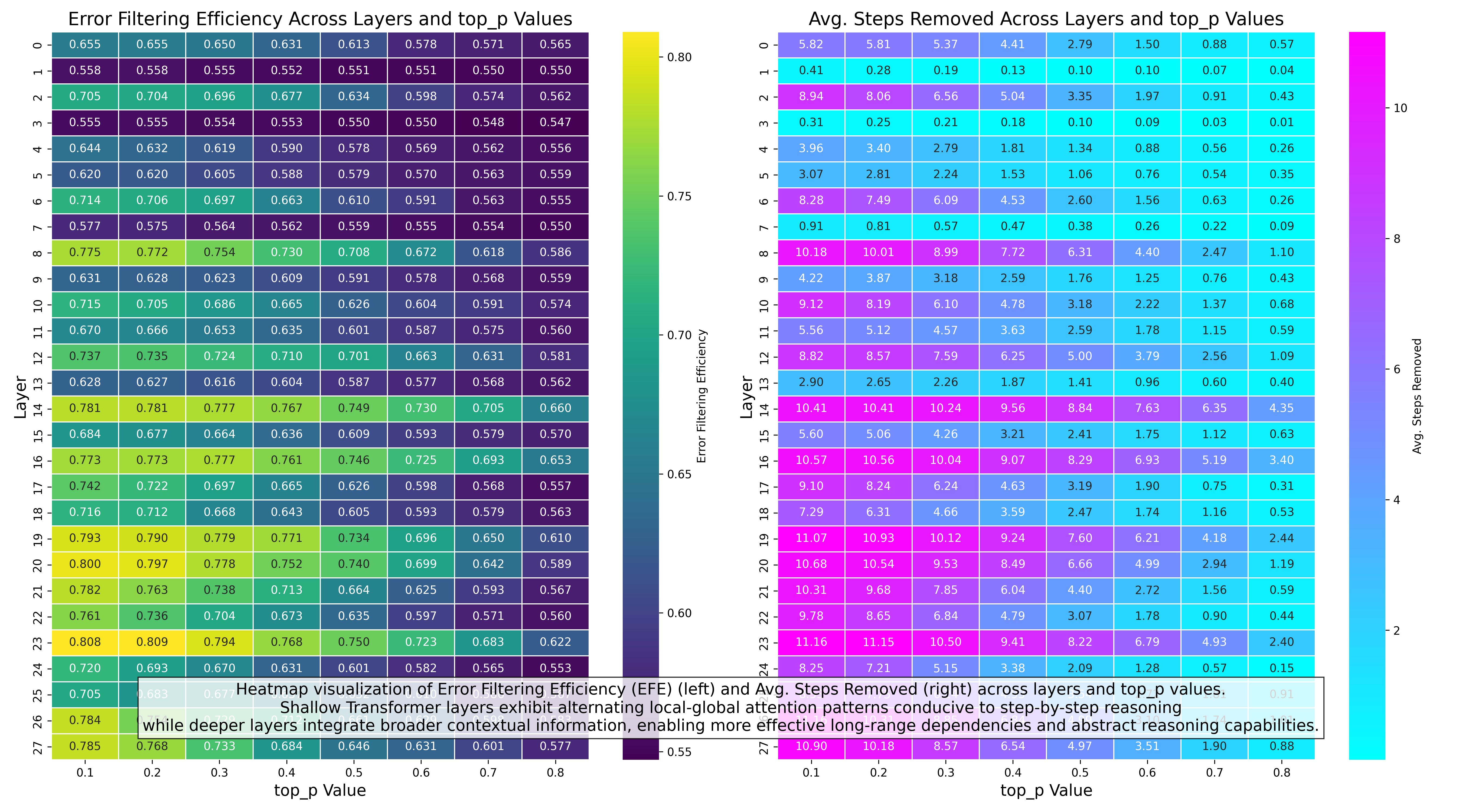}
        \caption{Attention heatmap visualization showing the alternating local-global focus pattern in early layers (0-13) versus consistent global attention in later layers (14-27).}
        \label{fig:layer_attention_heatmap}
    \end{subfigure}
    \caption{Layer-wise Analysis Using Step Matrix: Error Filtering Efficiency across Different Layers (Left) and Reasoning Step Intervals (Right) Based on Top-P Filtering Algorithm.}
    \label{fig:combined_layer_analysis}
\end{figure}

As shown in the Figure~\ref{fig:layer_attention_heatmap}, when using the step matrix of layer 0 to calculate the Top-P algorithm with a retention rate of 0.1, an average of 5.82 steps can be removed, indicating that the attention span of this layer spans at least more than 5 steps. However, under the same conditions, layer 1 can only remove 0.41 reasoning steps, suggesting that this layer's attention primarily focuses on the previous step. This alternating pattern consistently appears up to layer 13. When using the step matrix from layers 14-27 for the top-P algorithm with a retention rate of 0.1, the number of steps removed is consistently greater than 8, reaching a maximum of 11.06. This shows that later layers are interested in information from a wider range of reasoning steps. Similarly, we also found that these layers with broader attention spans can better describe the importance of reasoning steps (Figure~\ref{fig:best_layer}).
These findings may inspire future work on pruning reasoning models.

\paragraph{Analysis of Model Reasoning Patterns.}

In this section, we select 1k SR-FLOW model reasoning completions and construct a graph of directed edge relationship between the reasoning steps (Figure~\ref{fig:reasoning_path_graph}). We ignore edges with low frequency for clarity of presentation.
\begin{tcolorbox}[
  enhanced,
  title={\bfseries Takeaway 3},
  colback=white,
  colframe=yellow!70!black,
  colbacktitle=yellow!70!black,
  coltitle=white,
  boxrule=1pt,
  arc=4mm,
  fonttitle=\bfseries
]
During the problem-solving phase, various reasoning \textbf{loops} of different sizes form around steps like <verify>, depending on the difficulty of the problem.
\end{tcolorbox}

We find that when reasoning, SR-FLOW typically starts from problem definition (assumption), then proceeds through problem decomposition (decompose), moves to formalize, and finally reaches a conclusion through verify consequence. This represents a common trajectory. The paths involving assumption and summarize components are quite stable. However, we discovered multiple loop structures centered around verify: for example, verify-alternative, verify-caseAnalysis, and other complex loops. The model's different answers are often formed and corrected within these loops.

Figure~\ref{fig:tag_positions} shows the distribution of the relative positions of each tag within the reasoning process.


\section{Conclusion}
In this paper, we present a novel approach to enhance LLMs' structured reasoning capabilities. Our method combines cognitive science principles with neurosymbolic AI advances by explicitly incorporating structured knowledge representations and reasoning processes. The key contributions include: (1) a structured data transformation method with explicit reasoning step annotations, (2) two reinforcement learning algorithms, MAX-Flow for reasoning effectiveness evaluation, and LCS for computational efficiency, and (3) a lightweight fine-tuning strategy using only 500 examples and 250 RL steps. Our experiments on the DeepSeek-R1-Distill-Qwen-1.5B model demonstrate that this approach achieves more concise reasoning while maintaining robust performance across various scenarios. These results suggest that incorporating explicit structure into LLM reasoning processes can lead to more efficient and reliable AI systems.

\bibliographystyle{nips}  
\small
\bibliography{nips}
\normalsize


\appendix

\newpage
\section{Appendix}


\subsection{Part of Figures and Tables}
For better layout and presentation, we have placed some figures and tables in a unified location in the Appendix.

Table~\ref{tab:methods_comparison} presents a comprehensive comparison of three Small Structure Reasoning (SR) methods across various mathematical benchmarks. SR-FLOW demonstrates superior performance, achieving the highest average accuracy (58.1\%) while requiring fewer reasoning steps. SR-LCS offers the most token-efficient approach, using approximately 20\% fewer tokens while maintaining competitive accuracy. Highlighted cells indicate top performances for each benchmark and method, showing that different reasoning approaches excel in different problem domains.
\begin{table}[htbp]
\centering
\caption{Comparative Analysis of Reasoning Methods (Pass@1 Accuracy \& Efficiency)}
\label{tab:methods_comparison}
\resizebox{\textwidth}{!}{
\begin{tabular}{lccccccccc}
\toprule
\multirow{2}{*}{\textbf{Method}} & \multicolumn{7}{c}{\textbf{Accuracy (\%)}} & \textbf{Steps} & \textbf{Tokens} \\
\cmidrule(lr){2-8}
  & AIME24 & AIME25 & AMC23 & MATH500 & Minerva & Olympiad & Large Avg.& Avg. & Avg. \\
\midrule
\textbf{SR-ACC} 
 & 32.7±8.7 & 25.3±8.0 & 75.8±6.7 & 85.6±1.6 & 31.3±2.8 & 53.3±1.9 & 56.7±2.1 & 9.57 & 1873 \\
 & 36.7±8.9 & 26.7±8.2 & 72.0±7.1 & 85.5±1.6 & 31.1±2.8 & 53.3±1.9 & 56.6±2.1 & 9.94 & 1796 \\
 & 30.0±8.5 & 21.3±7.6 & 74.0±7.0 & 84.7±1.6 & 32.4±2.8 & 52.9±1.9 & 56.7±2.1 & 10.41 & 1808 \\
 & 24.0±7.8 & 22.5±7.7 & 74.0±7.0 & 83.7±1.7 & 33.0±2.9 & 52.2±1.9 & 56.3±2.2 & 10.81 & 1859 \\
 & 30.3±8.7 & 24.2±7.8 & 73.5±6.6 & 84.2±1.6 & 31.7±2.8 & 50.8±1.9 & 55.6±2.1 & 11.03 & 1828 \\
 & 36.7±7.7 & 19.5±7.2 & 70.6±7.4 & 83.7±1.7 & 31.1±2.8 & 50.9±1.9 & 55.2±2.1 & 10.85 & 1854 \\
\addlinespace
\textbf{SR-FLOW} 
 & 32.7±8.7 & 25.3±8.0 & 75.8±6.7 & 85.6±1.6 & 31.3±2.8 & 53.3±1.9 & 56.7±2.1 & 9.57 & 1873 \\
 & 33.0±8.5 & 26.0±8.1 & 76.5±6.8 & 84.9±1.6 & 31.3±2.8 & 53.6±1.9 & 56.6±2.1 & 9.63 & 1820 \\
 & 34.7±8.3 & 26.3±7.8 & 76.8±6.8 & 85.6±1.6 & 34.1±2.9 & 53.5±1.9 & 57.7±2.1 & 9.52 & 1779 \\
 & 36.7±8.9 & 27.0±8.2 & 77.8±6.6 & 85.3±1.6 & 34.2±2.9 & 54.8±1.9 & 58.1±2.1 & 9.24 & 1830 \\
 & 33.5±8.2 & 25.7±7.6 & 75.3±6.9 & 85.0±1.6 & 33.2±2.9 & 53.7±1.9 & 57.3±2.1 & 8.78 & 1804 \\
 & 30.3±8.7 & 24.2±7.8 & 74.0±7.0 & 84.7±1.6 & 32.3±2.8 & 54.3±1.9 & 57.1±2.1 & 7.84 & 1798 \\
\addlinespace
\textbf{SR-LCS} 
 & 32.7±8.7 & 25.3±8.0 & 75.8±6.7 & 85.6±1.6 & 31.3±2.8 & 53.3±1.9 & 56.7±2.1 & 9.57 & 1873 \\
 & 34.0±8.7 & 24.7±7.9 & 75.0±6.9 & 84.9±1.6 & 32.0±2.8 & 54.1±1.9 & 57.0±2.1 & 9.61 & 1780 \\
 & 33.3±8.7 & 22.5±7.8 & 74.0±7.0 & 84.4±1.6 & 30.9±2.8 & 51.8±1.9 & 55.7±2.1 & 10.36 & 1668 \\
 & 30.3±8.5 & 23.3±7.8 & 74.0±7.0 & 83.3±1.7 & 29.9±2.8 & 51.6±1.9 & 54.9±2.1 & 10.91 & 1614 \\
 & 33.7±8.7 & 23.7±7.9 & 75.0±6.9 & 84.9±1.6 & 31.8±2.8 & 50.6±1.9 & 55.8±2.1 & 11.75 & 1509 \\
 & 31.0±8.5 & 23.3±7.7 & 74.8±7.1 & 84.8±1.6 & 30.5±2.8 & 52.7±1.9 & 56.0±2.1 & 11.56 & 1504 \\
\bottomrule
\end{tabular}
}
This table demonstrates the evolution of various metrics during GRPO training from 0 to 250 global steps under three reward functions: ACC, FLOW, and LCS. The metrics are tracked throughout the training process to show how different reward mechanisms influence model performance.
\end{table}

\begin{figure}[htbp]
    \centering
    \begin{subfigure}[b]{0.48\textwidth}
        \centering
        \includegraphics[width=\linewidth]{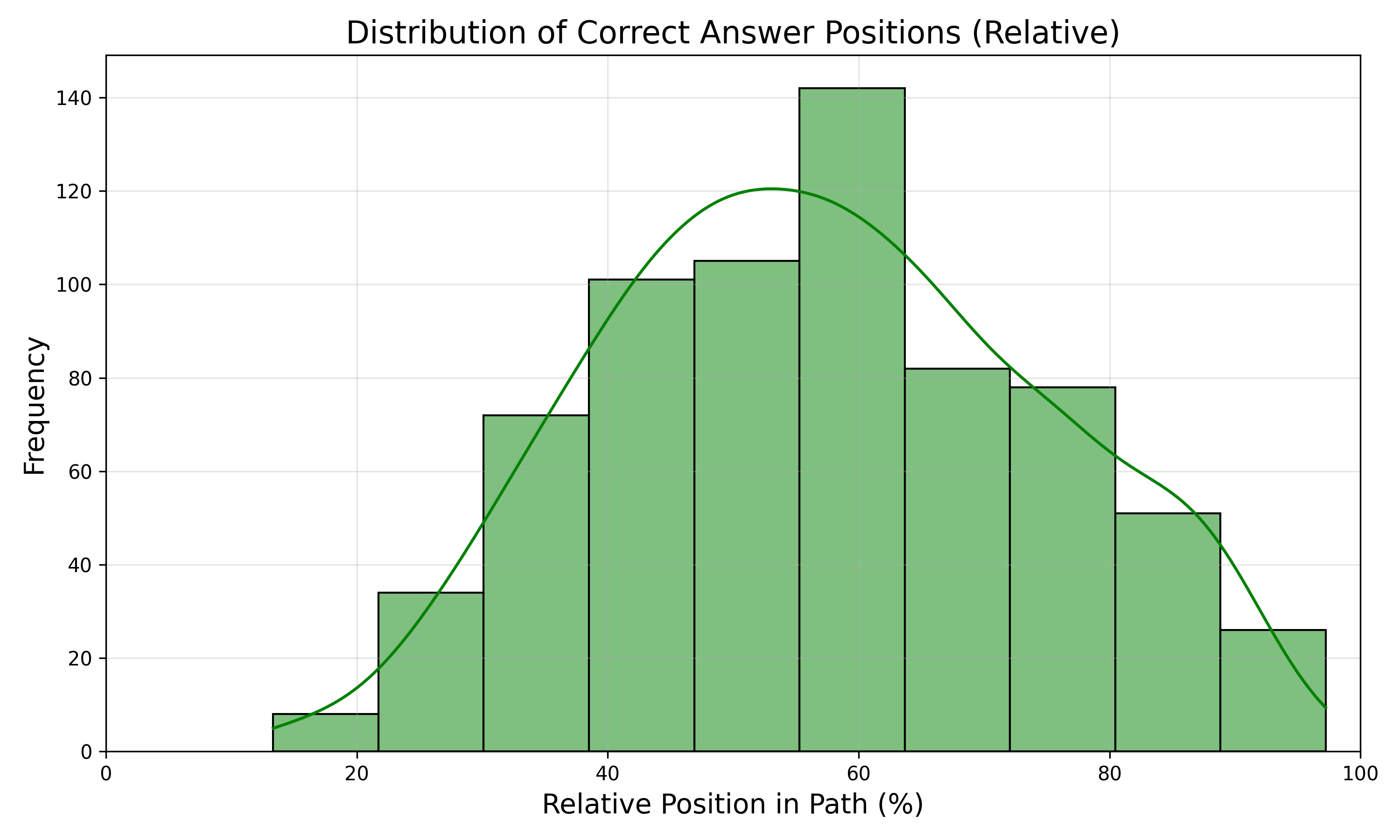}
        \caption{First Correct Answer Position Distribution}
        \label{fig:answer_pos_sub}
    \end{subfigure}
    \hfill
    \begin{subfigure}[b]{0.48\textwidth}
        \centering
        \includegraphics[width=\linewidth]{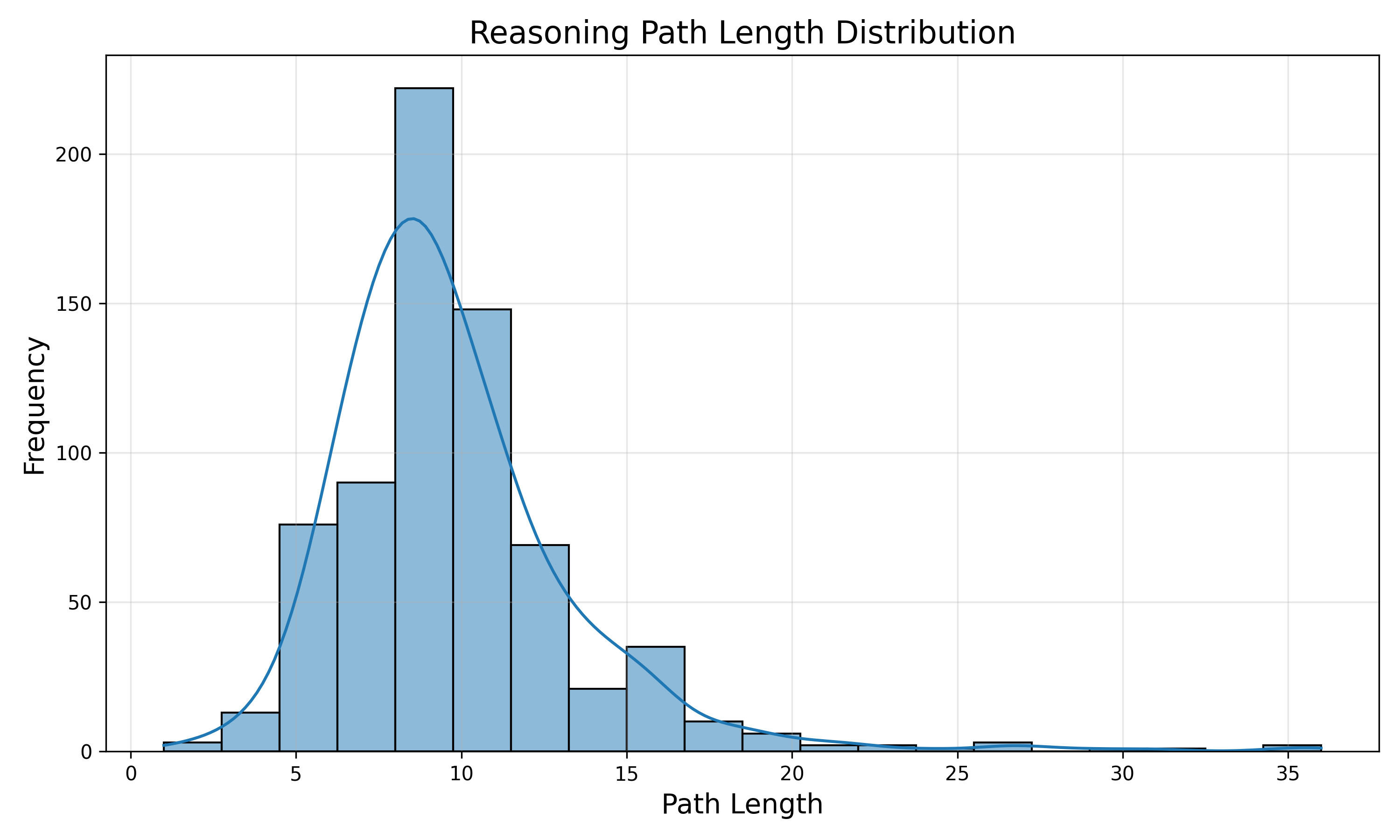}
        \caption{Statistical distribution of reasoning path lengths}
        \label{fig:path_length_sub}
    \end{subfigure}
    \caption{Analysis of Model Reasoning Patterns: Distribution of First Correct Answers (Left) and Reasoning Path Lengths (Right).}
    \label{fig:reasoning_analysis}
\end{figure}

\begin{figure}[ht]
\centering
\begin{subfigure}[b]{0.495\textwidth}
    \centering
    \includegraphics[width=\linewidth]{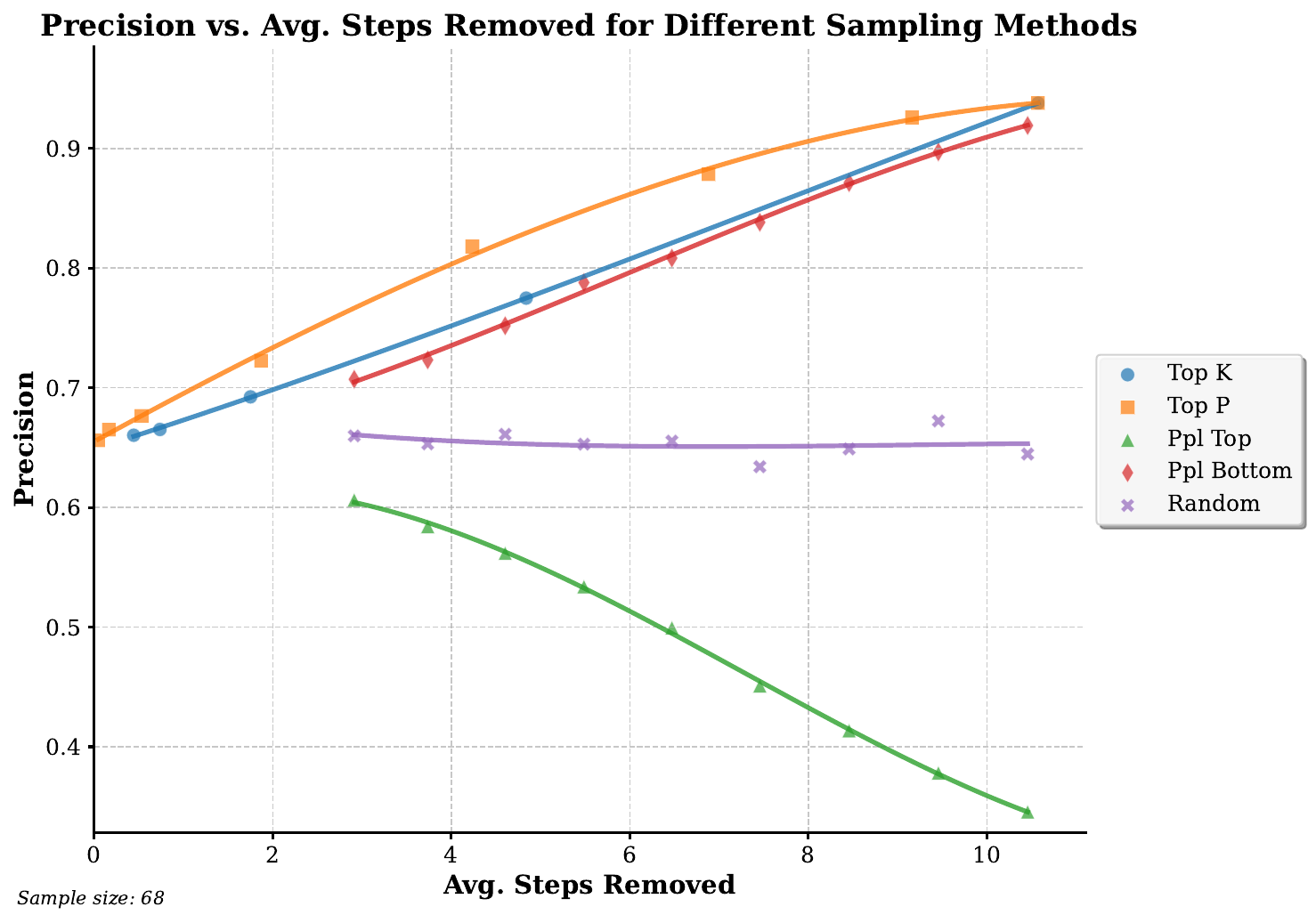}
    \caption{Redundant but harmless information}
    \label{fig:message_not_useful}
\end{subfigure}
\hfill
\begin{subfigure}[b]{0.495\textwidth}
    \centering
    \includegraphics[width=\linewidth]{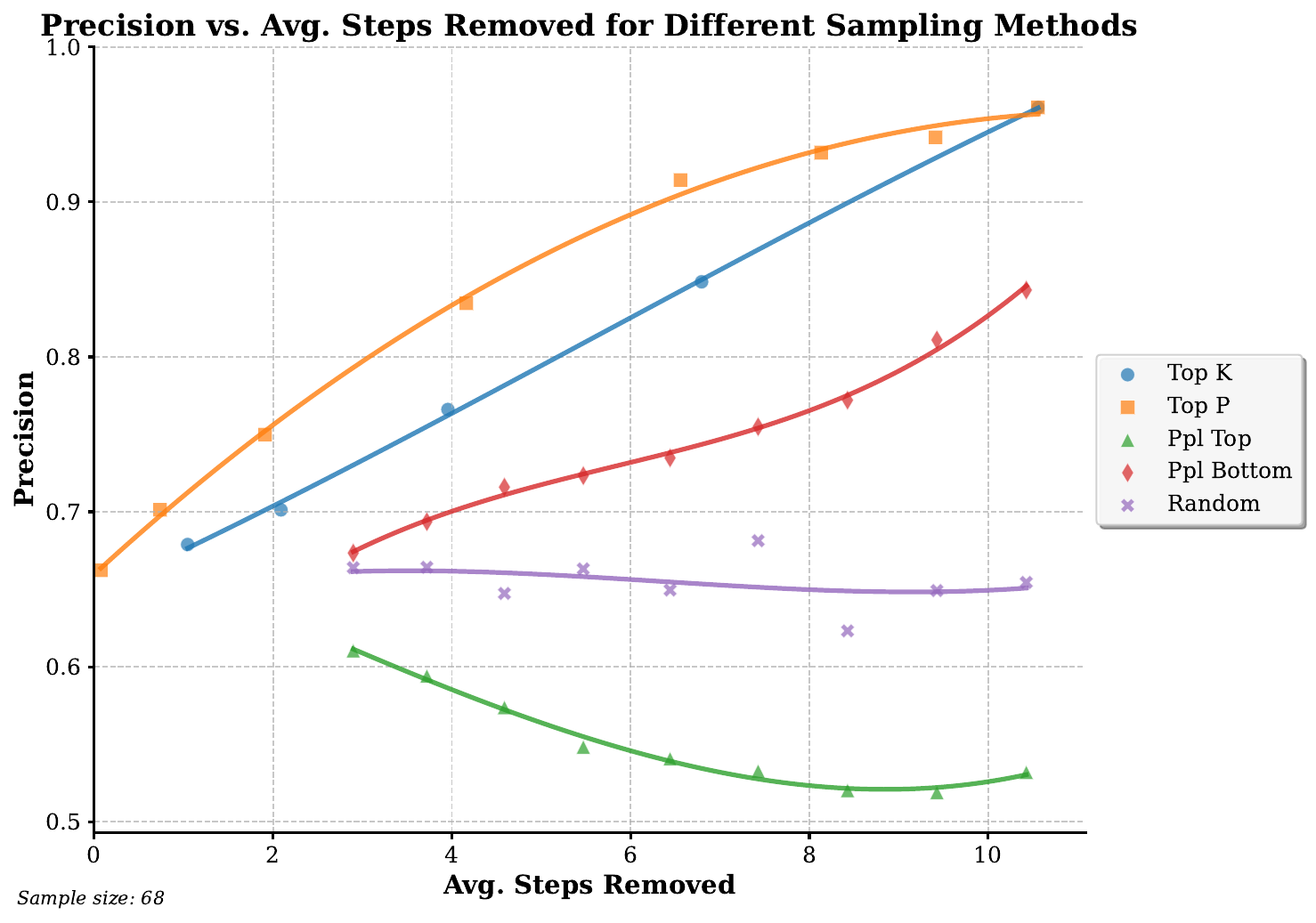}
    \caption{Distracted but harmless information}
    \label{fig:mind_wandering}
\end{subfigure}

\begin{subfigure}[b]{0.495\textwidth}
    \centering
    \includegraphics[width=\linewidth]{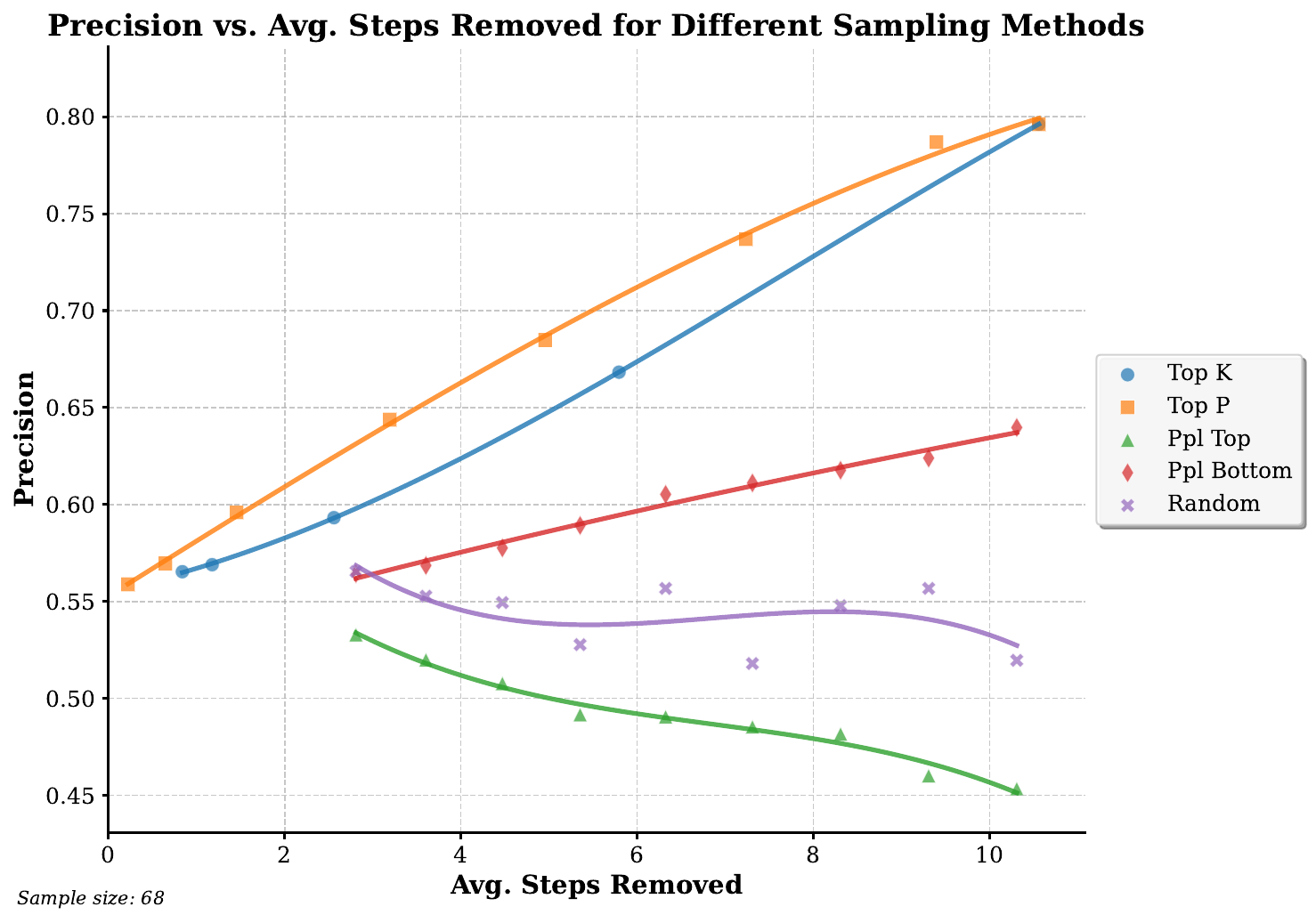}
    \caption{Harmful irrelevant reasoning}
    \label{fig:mix_intervene}
\end{subfigure}
\hfill
\begin{subfigure}[b]{0.495\textwidth}
    \centering
    \includegraphics[width=\linewidth]{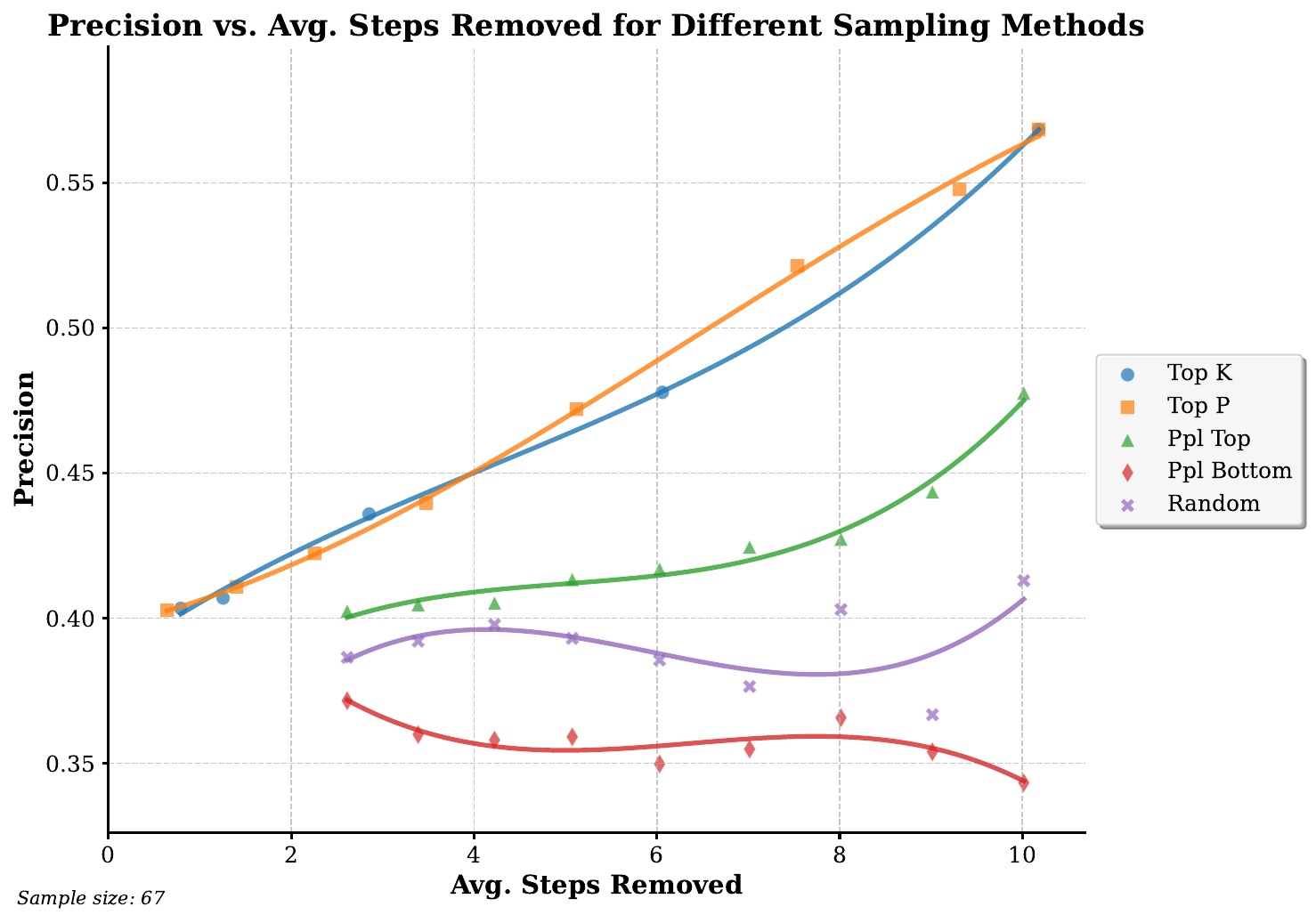}
    \caption{Logically confused information}
    \label{fig:same_meaning}
\end{subfigure}
\caption{IISR Results: Error Filtering Efficiency (Precision) of different algorithms when removing 1-11 steps under four types of information interference.}
\label{fig:EFE_comparision}
\end{figure}

\begin{figure}[ht]
\centering
\includegraphics[width=1.0\textwidth]{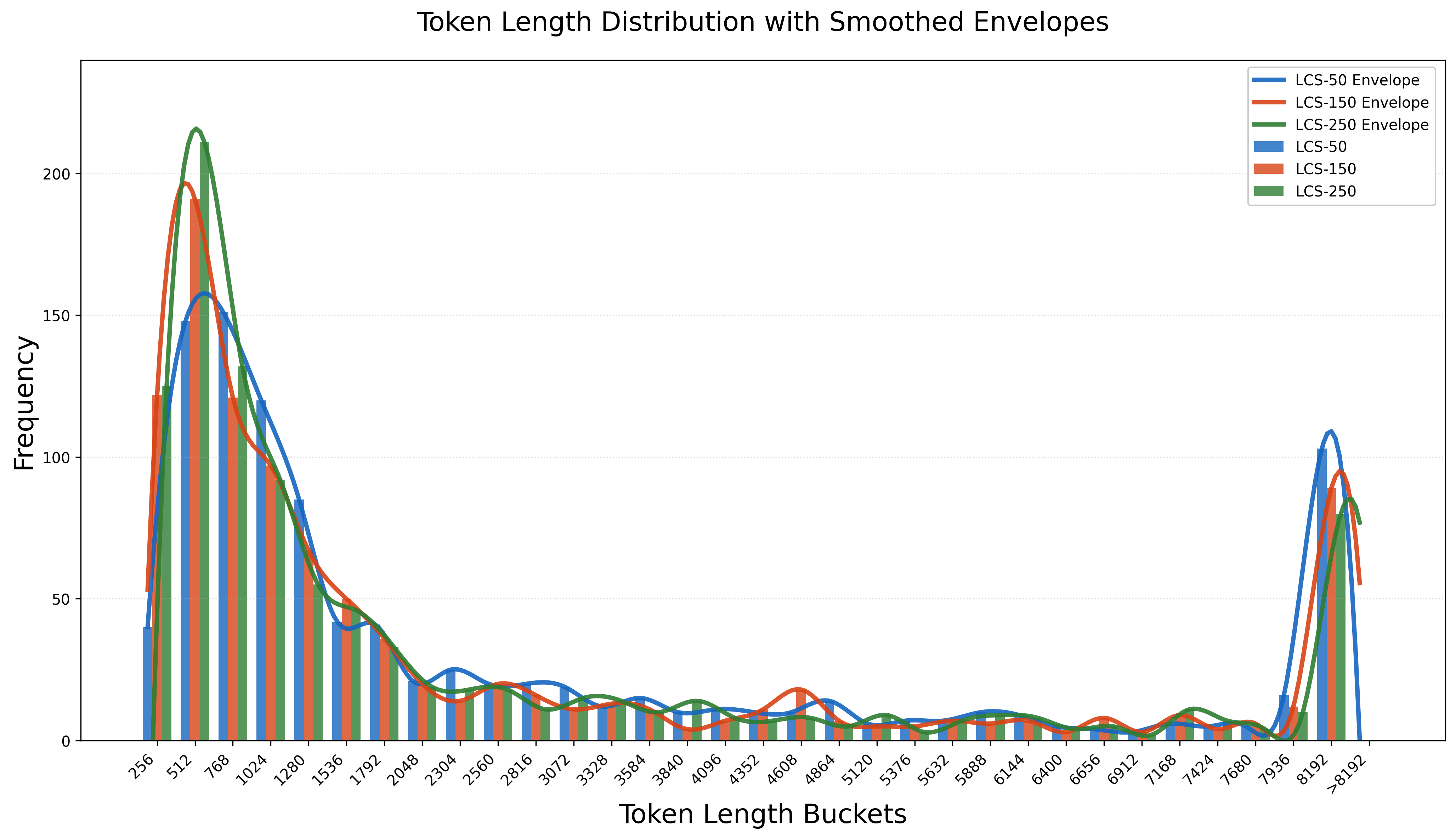}
\caption{Token length distribution of LCS models under different training stages. The smoothed envelopes show how reward training shifts the distribution towards optimal reasoning lengths (512-1024 tokens) while maintaining performance.}
\label{fig:lcs_dist}
\end{figure}


\begin{table}[h]
\centering
\caption{Early Stopping Detection Parameters and Sample Statistics}
\resizebox{\columnwidth}{!}{
\begin{tabular}{lc}
\hline
\textbf{Metric} & \textbf{Value} \\
\hline
Number of valid samples & 705 \\
Top five frequent tags & verify, summarize, equivalent, formalize, consequence \\
Token interval & 128 \\
Top five useful words & but, wait, however, check, alternatively \\
\hline
\end{tabular}
}
\label{tab:trigger_stats}
\end{table}

\subsection{Full Prompts}
\label{full_prompts}
\begin{tcolorbox}[
  enhanced,
  title={\bfseries Mathematical Problem Solving Template},
  colback=white,
  colframe=blue!70!black,
  colbacktitle=blue!70!black,
  coltitle=white,
  boxrule=1pt,
  arc=4mm,
  fonttitle=\bfseries
]
Please use the following tags at the beginning of each sentence in your reasoning: <rephrase>, <inference>, <analogy>, <equivalent>, <association>, <reverse>, <summarize>, <verify>, <complete>, <decompose>, <counterexample>, <assumption>, <constraint>, <case\_analysis>, <contradiction>, <abstraction>, <formalize>, <generalize>, <specialize>, <critique>, <alternative>, <consequence>, <intuition>.

\{Question\}

Please reason step by step, and put your final answer within boxed\{ \}.
\end{tcolorbox}

\begin{tcolorbox}[
  enhanced,
  title={\bfseries Multiple Choice Problem Template},
  colback=white,
  colframe=green!70!black,
  colbacktitle=green!70!black,
  coltitle=white,
  boxrule=1pt,
  arc=4mm,
  fonttitle=\bfseries
]
Please use the following tags at the beginning of each sentence in your reasoning: <rephrase>, <inference>, <analogy>, <equivalent>, <association>, <reverse>, <summarize>, <verify>, <complete>, <decompose>, <counterexample>, <assumption>, <constraint>, <case\_analysis>, <contradiction>, <abstraction>, <formalize>, <generalize>, <specialize>, <critique>, <alternative>, <consequence>, <intuition>.

\{Question\}

A) \{A\} \\
B) \{B\} \\
C) \{C\} \\
D) \{D\} 

Please reason step by step, and answer the following multiple choice question. The last line of your response should be of the following format: 'Answer: \$LETTER' (without quotes) where LETTER is one of ABCD.
\end{tcolorbox}

\subsection{Other Step Importance Evaluation Algorithm Implementation}
\label{other algorithms}
\textbf{Top-P and Top-K Selection.}
Based on the step matrix computed from different layers (See Section~\ref{method:step-matrix}), we implement backtracking selection methods:

\begin{equation}
    \text{SelectSteps}(A, k, p) = \{s_i\}_{i=0}^{m},
\end{equation}

where $A \in \mathbb{R}^{n \times n}$ is the step attention matrix, and we select steps starting from the last step $s_{n-1}$ by either: 
Top-K: For each step $s_i$, select up to $k$ preceding steps with highest attention scores.
Top-P: Select preceding steps with cumulative normalized attention exceeding threshold $p$.

The algorithm traverses backward from the final step, adding important preceding steps to a visited set based on attention weights, ensuring all critical reasoning dependencies are captured.
\textbf{Average Perplexity.}
For each step, we compute token-level perplexity:
\begin{equation}
    \text{Perplexity}(t_i) = \frac{1}{P(t_i|x,t_1,...,t_{i-1})},
\end{equation}

where $P(t_i|x,t_1,...,t_{i-1})$ is the probability of token $t_i$ given the prompt $x$ and all preceding tokens, derived from the softmax of logits:

\begin{equation}
    P(t_i|x,t_1,...,t_{i-1}) = \frac{\exp(\text{logits}_i)}{\sum_j \exp(\text{logits}_j)}.
\end{equation}

The average perplexity for a step $s$ containing tokens $\{t_1, t_2, ..., t_m\}$ is:

\begin{equation}
    \text{AvgPerplexity}(s) = \exp\left(-\frac{1}{m}\sum_{i=1}^{m}\log P(t_i|x,t_1,...,t_{i-1})\right).
\end{equation}


\textbf{Random Selection.}
A baseline approach where steps are selected randomly without leveraging attention patterns or perplexity metrics.

\begin{figure}[h]
    \centering
    \includegraphics[width=1.0\textwidth]{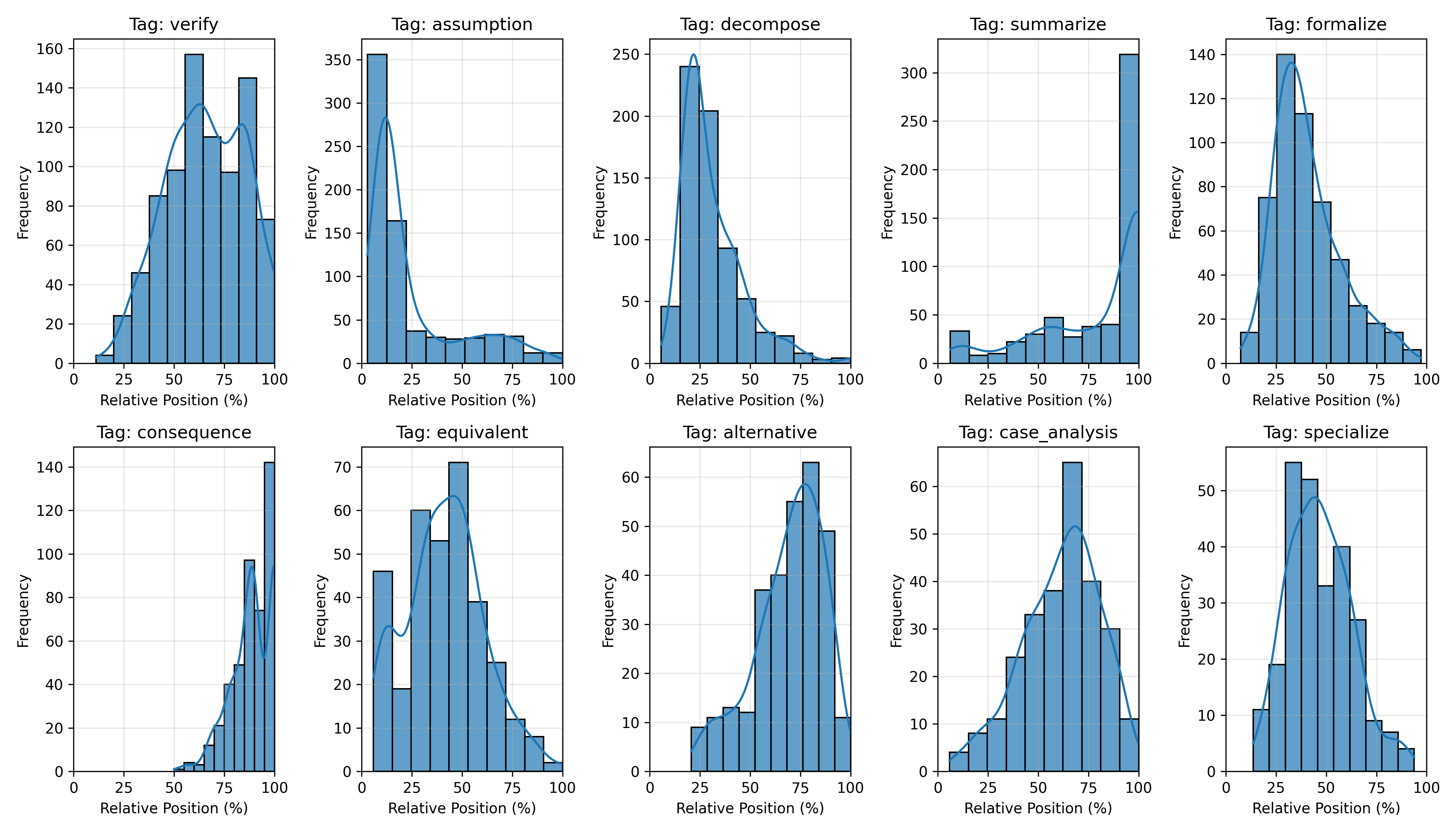}
    \caption{Distribution of the relative positions of each tag within the reasoning process.}
    \label{fig:tag_positions}
\end{figure}

\begin{figure}[ht]
    \centering
    \includegraphics[width=1.0\textwidth]{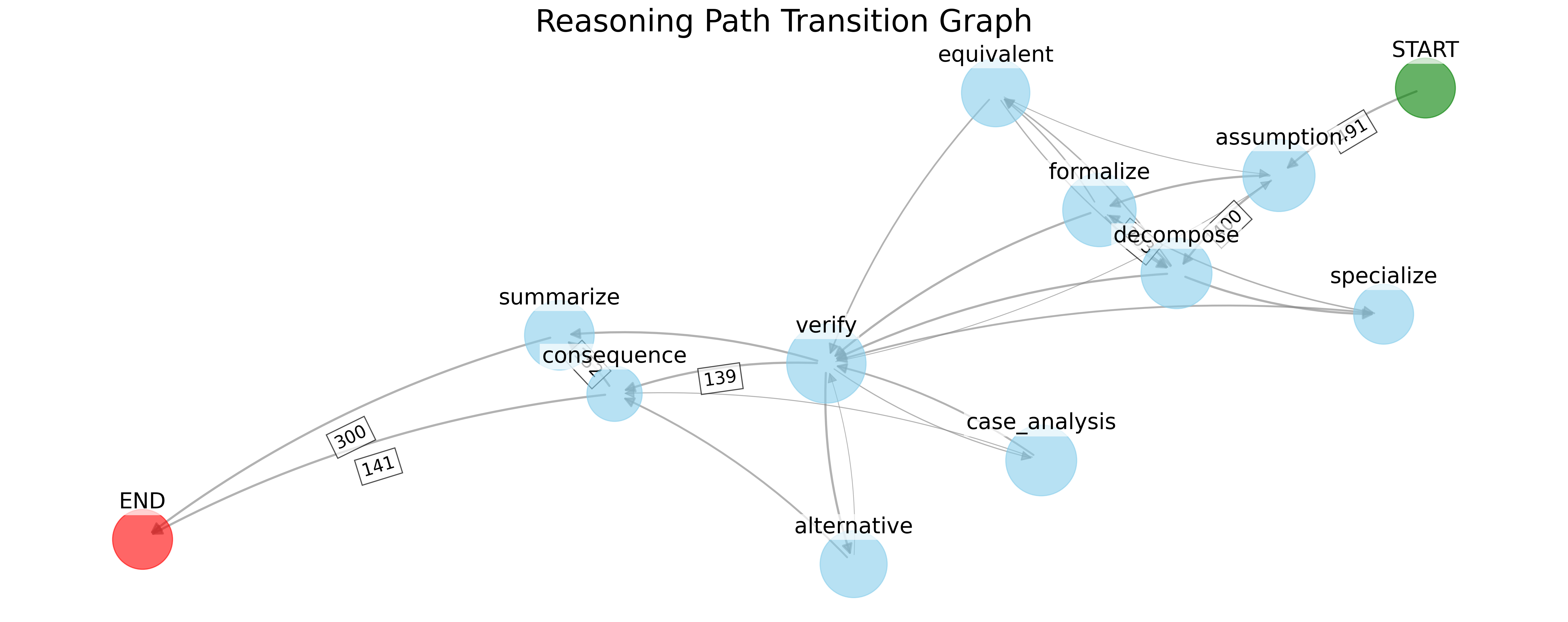}
    \caption{Illustration of Reasoning Path Transition Graph.}
    \label{fig:reasoning_path_graph}
\end{figure}

\subsection{Error Filtering Efficiency (EFE) Evaluation Formula}
\begin{figure}[ht]
    \centering
    \includegraphics[width=1\textwidth]{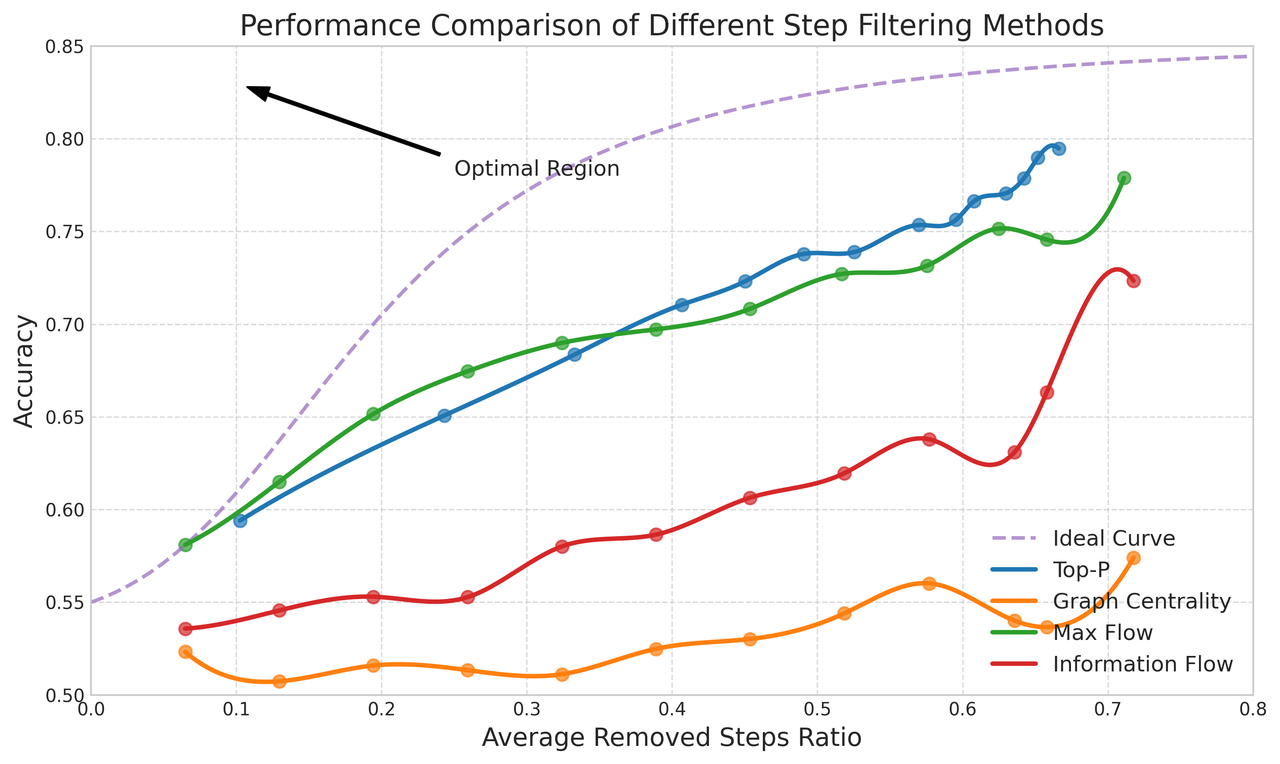}
    \caption{Comparison of Algorithms for Error Filtering Efficiency Averaged Across Four Tasks.}
    \label{fig:method_comparison}
\end{figure}
\label{efe formula}
For the IISR experiment, where we randomly inject $N$ interference steps into an $M$-step reasoning process, the Error Filtering Efficiency is calculated as:

\begin{equation}
    \text{EFE} = 1 - \frac{|\text{RetainedIrrelevantSteps}|}{|\text{IrrelevantSteps}|},
\end{equation}

where $|\text{IrrelevantSteps}|$ is the total number of interference steps injected ($N$), $|\text{RetainedIrrelevantSteps}|$ is the number of interference steps that were incorrectly retained after filtering

EFE measures the algorithm's ability to identify and remove irrelevant steps, with a value of 1.0 indicating perfect filtering (all interference steps removed) and 0.0 indicating no filtering capability.

As shown in Figure~\ref{fig:EFE_comparision}, we first compared Top K, Top P, Ppl Top (where higher perplexity indicates higher step importance), Ppl Bottom (the opposite), and Random. We evaluated the Error Filtering Efficiency under four different interference injection methods. The results show that Ppl-based methods exhibit unstable performance across different tasks.

As illustrated in Figure~\ref{fig:method_comparison}, we further compared the better performing methods: Top-P, Max-Flow, and Information-Flow. We found that the Max Flow method demonstrates a superior ability in evaluating reasoning steps, particularly when removing a small number of steps.

\newpage
\subsection{Structure Reasoning Process Demonstration}
\label{reasoning example}
\definecolor{methodcolor}{RGB}{0,100,0}
\definecolor{insertcolor}{RGB}{178,34,34}
\definecolor{tagcolor}{RGB}{0,0,128}
\definecolor{promptcolor}{RGB}{106,13,173}
\definecolor{iocolor}{RGB}{0,100,0}
\newcommand{\methodtext}[1]{\textcolor{methodcolor}{\small\textsf{#1}}}
\newcommand{\io}[1]{\textcolor{iocolor}{\small\textsf{#1}}}
\newcommand{\thoughttag}[1]{\textcolor{tagcolor}{\small\textbf{<#1>}}}
\newcommand{\inserttext}[1]{\textcolor{insertcolor}{\small\textit{#1}}}
\newcommand{\prompttext}[1]{\textcolor{promptcolor}{\small\textsf{#1}}{\small\textbf{$\langle$#1$\rangle$}}}
\newmdenv[
  skipabove=1em,
  skipbelow=1em,
  linewidth=0.5pt,
  linecolor=gray,
  backgroundcolor=gray!5
]{reasoningbox}

\begin{figure}[h]
\centering
\includegraphics[width=1.\textwidth]{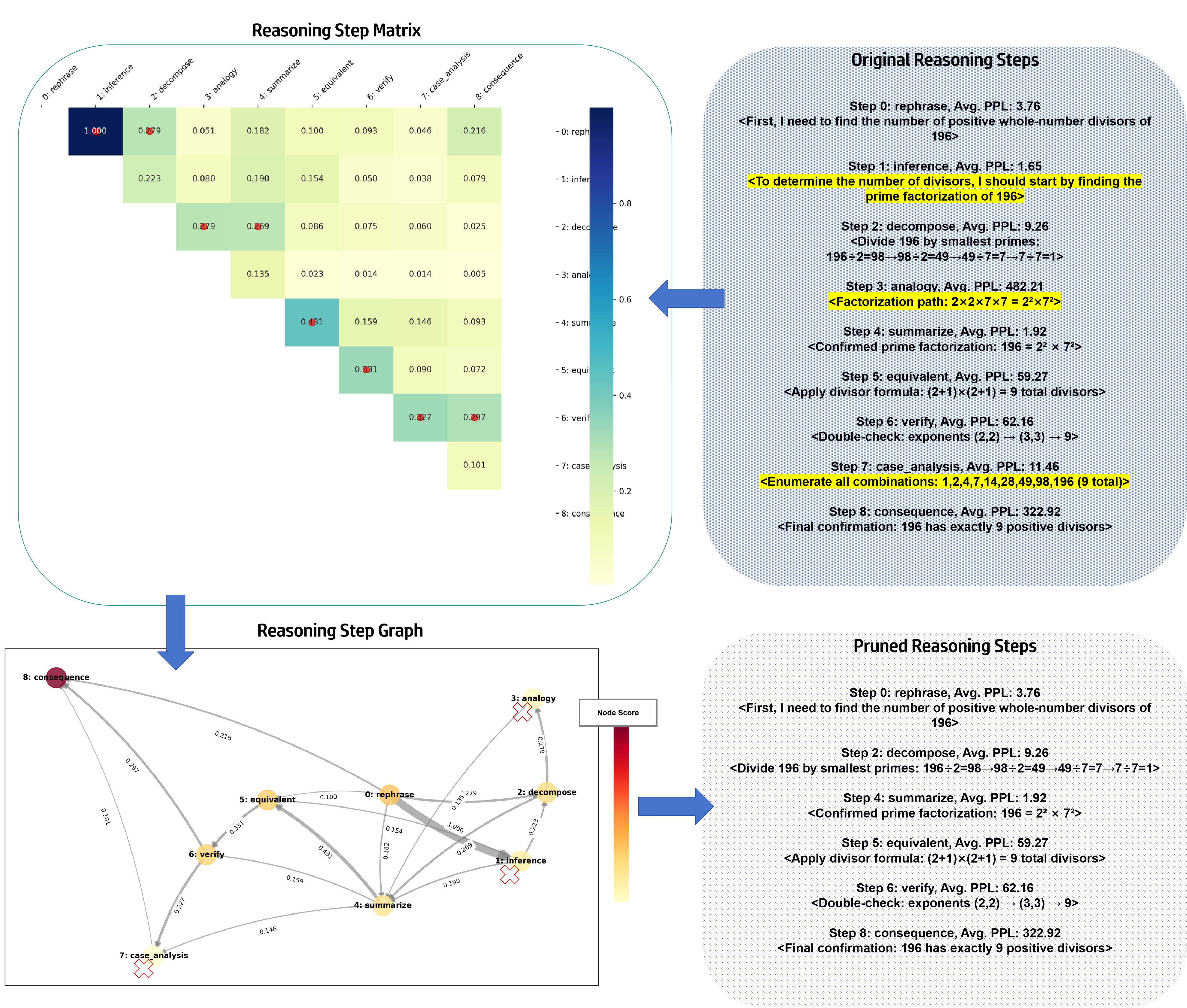}
\caption{Using Max-flow Method for Filtering Steps during IISR Experiment.}
\label{fig:dreaming_test}
\end{figure}

\io{\textbf{Input Prompt:}}
\begin{quote}
Please use the following tags at the beginning of each sentence in your reasoning:
<rephrase>, <inference>, <analogy>, <equivalent>, <association>, <reverse>, <summarize>, <verify>, <complete>, <decompose>, <counterexample>, <assumption>, <constraint>, <case\_analysis>, <contradiction>, <abstraction>, <formalize>, <generalize>, <specialize>, <critique>, <alternative>, <consequence>, <intuition> \\[0.5em]
Convert the point $(0,3)$ in rectangular coordinates to polar coordinates. Enter your answer in the form $(r,\theta),$ where $r > 0$ and $0 \le \theta < 2 \pi.$ \\[0.5em]
Please reason step by step, and put your final answer within \boxed{ }.
\end{quote}

\begin{reasoningbox}
\thoughttag{summarize} 
The task is to convert the rectangular coordinates $(0, 3)$ to polar coordinates $(r, \theta)$ with $r > 0$ and $0 \leq \theta < 2\pi$.

\thoughttag{assumption} 
In rectangular (Cartesian) coordinates, a point is represented as $(x, y)$. To convert to polar coordinates $(r, \theta)$, we need to find:
\begin{itemize}[leftmargin=*]
  \item $r$: distance from origin
  \item $\theta$: angle with positive x-axis
\end{itemize}

\thoughttag{formalize} 
Conversion formulas:
\begin{align*}
r &= \sqrt{x^2 + y^2} \\
\theta &= \arctan(y/x) \quad \text{(when $x \neq 0$)}
\end{align*}

\thoughttag{decompose} 
Given point $(0, 3)$:
\begin{itemize}[leftmargin=*]
  \item $x = 0$
  \item $y = 3$
\end{itemize}

\thoughttag{specialize} 
Calculate $r$:
\[ r = \sqrt{0^2 + 3^2} = 3 \]

\thoughttag{reverse} 
For $\theta$: $\arctan(3/0)$ is undefined. Observation:
\begin{itemize}[leftmargin=*]
  \item $x = 0$ and $y > 0$ $\Rightarrow$ point lies on positive y-axis
\end{itemize}

\thoughttag{case\_analysis} 
Special case when $x = 0$:
\[ \theta = \begin{cases}
   \pi/2 & \text{if } y > 0 \\
   3\pi/2 & \text{if } y < 0
\end{cases} \]

\thoughttag{verify} 
Check $\theta$ range: $\pi/2 \in [0, 2\pi)$ is valid.

\thoughttag{complete} 
Final polar coordinates:
\[ \boxed{\left(3, \frac{\pi}{2}\right)} \]
\end{reasoningbox}

\io{\textbf{Output:}}
\begin{quote}
The polar coordinates are \boxed{\left(3, \frac{\pi}{2}\right)}.
\end{quote}

\io{\textbf{Input Prompt:}}
\begin{quote}
Please use the following tags at the beginning of each sentence in your reasoning:
<rephrase>, <inference>, <analogy>, <equivalent>, <association>, <reverse>, <summarize>, <verify>, <complete>, <decompose>, <counterexample>, <assumption>, <constraint>, <case\_analysis>, <contradiction>, <abstraction>, <formalize>, <generalize>, <specialize>, <critique>, <alternative>, <consequence>, <intuition> \\[0.5em]
What is $\sqrt{53}$ in simplest radical form? \\[0.5em]
Please reason step by step, and put your final answer within \boxed{ }.
\end{quote}

\begin{reasoningbox}
\thoughttag{assumption} 
To find the simplest radical form of $\sqrt{53}$, I need to check if 53 has any perfect square factors.

\thoughttag{decompose} 
First, let's factorize 53 to see its prime components. Since 53 is a prime number, its only factors are 1 and 53.

\thoughttag{verify} 
Since 53 is prime, there are no smaller factors other than 1 and itself. Therefore, $\sqrt{53}$ cannot be simplified further because there are no perfect squares greater than 1 that divide 53.

\thoughttag{conclusion} 
Thus, the simplest radical form of $\sqrt{53}$ is $\sqrt{53}$ itself.
\end{reasoningbox}

\io{\textbf{Output:}}
\begin{quote}
The simplest radical form is \boxed{\sqrt{53}}.
\end{quote}

\subsection{Example Demonstration of Interference Injection and Selective Removal (\textbf{IISR})}
\label{ablation:IISR}
\begin{table}[htbp]
\centering
\caption{Steps retained by different filtering methods}
\label{tab:retention}
\begin{tabular}{l>{\ttfamily}l}
\toprule
\textbf{Method} & \textbf{Retained Steps} \\
\midrule
Top-P (p=0.7) & [1, 3, 5, 10, 11] \\
Maximum Flow & [1, 3, 5, 7, 10, 11] \\
Graph Centrality & [3, 4, 6, 8, 9, 10, 11] \\
Information Flow & [1, 2, 3, 4, 5, 6, 7, 8, 10, 11] \\
\bottomrule
\end{tabular}
\end{table}

\begin{itemize}[leftmargin=*]
\item Steps marked in red: [2, 4, 6, 9]
\item Interference types:
  \begin{itemize}
  \item Irrelevant pauses (Step 2)
  \item Off-topic reasoning (Steps 4)
  \item Disordered repetition (Steps 6)
  \item Redundant Reasoning (Step 9)
  \end{itemize}
\end{itemize}

\io{\textbf{Input Prompt:}}
\begin{quote}
Please use the following tags at the beginning of each sentence in your reasoning:
<rephrase>, <inference>, <analogy>, <equivalent>, <association>, <reverse>, <summarize>, <verify>, <complete>, <decompose>, <counterexample>, <assumption>, <constraint>, <case\_analysis>, <contradiction>, <abstraction>, <formalize>, <generalize>, <specialize>, <critique>, <alternative>, <consequence>, <intuition> \\[0.5em]
What is the smallest positive perfect cube that can be written as the sum of three?\\[0.5em]
Please reason step by step, and put your final answer within \boxed{ }.
\end{quote}

\begin{reasoningbox}
\thoughttag{rephrase} 
The problem is to find the smallest positive perfect cube expressible as the sum of three consecutive integers.

\inserttext{\thoughttag{inference} 
[Interference Step] Pausing to consider the implications of the next step.}

\thoughttag{inference} 
First, recall that perfect cubes are numbers like 1, 8, 27. The sum of three consecutive integers can be represented as $3n$.

\inserttext{\thoughttag{inference} 
[Off-topic Interference] Gravity cancels out, so mass and distances are the key variables...}

\thoughttag{analogy} 
Let the integers be $(n-1, n, n+1)$, then sum is $3n$. The smallest cube divisible by 3 is 27.

\inserttext{\thoughttag{inference} 
[Disordered Reasoning] Smaller cubes fail: $1=1^3$ gives non-integer $n=\frac{1}{3}$, $8=2^3$ gives $n=\frac{8}{3}$.}

\thoughttag{verify} 
Check 27: $8+9+10=27$. Is there a smaller cube? No, since 1 and 8 aren't divisible by 3.

\thoughttag{assumption} 
Alternative representation: $(n, n+1, n+2)$ sums to $3(n+1)$. Still requires divisibility by 3.

\inserttext{\thoughttag{reverse} 
[Redundant Reasoning] Alternatively, maybe the three consecutive integers could be...}

\thoughttag{consequence} 
Smaller cubes fail: $1=1^3$ gives non-integer $n=\frac{1}{3}$, $8=2^3$ gives $n=\frac{8}{3}$.

\thoughttag{summarize} 
Therefore, the smallest positive perfect cube meeting the criteria is:
\[ \boxed{27} \]
\end{reasoningbox}

\subsection{Experimental Details}
\label{experimental_setting}

\textbf{ Truncation Robustness.}

\begin{figure}[h]
\centering
\begin{subfigure}[b]{0.495\textwidth}
    \centering
    \includegraphics[width=\linewidth]{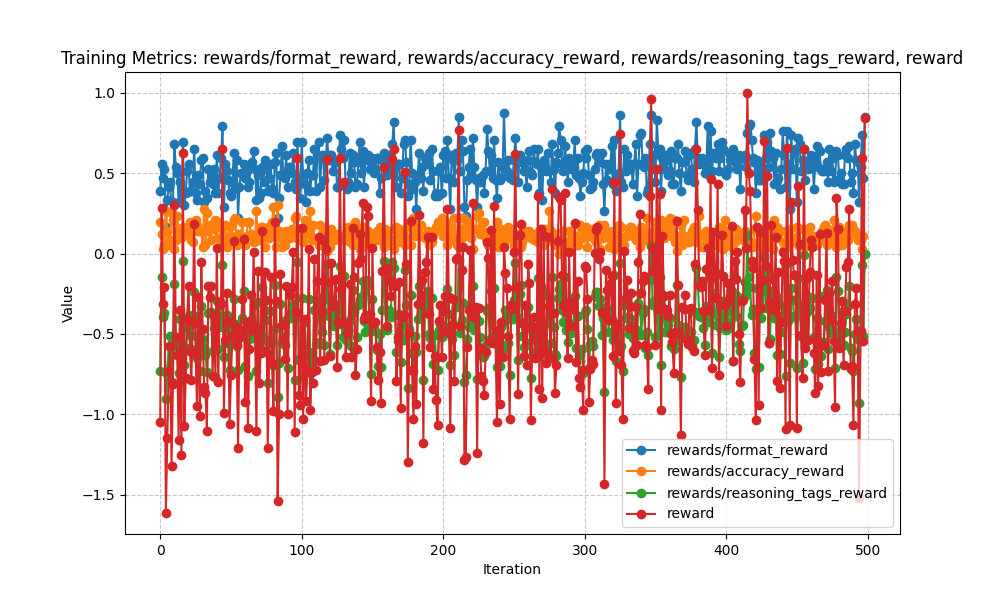}
    \caption{Reward curve with masking}
    \label{fig:w_mask_reward}
\end{subfigure}
\hfill
\begin{subfigure}[b]{0.495\textwidth}
    \centering
    \includegraphics[width=\linewidth]{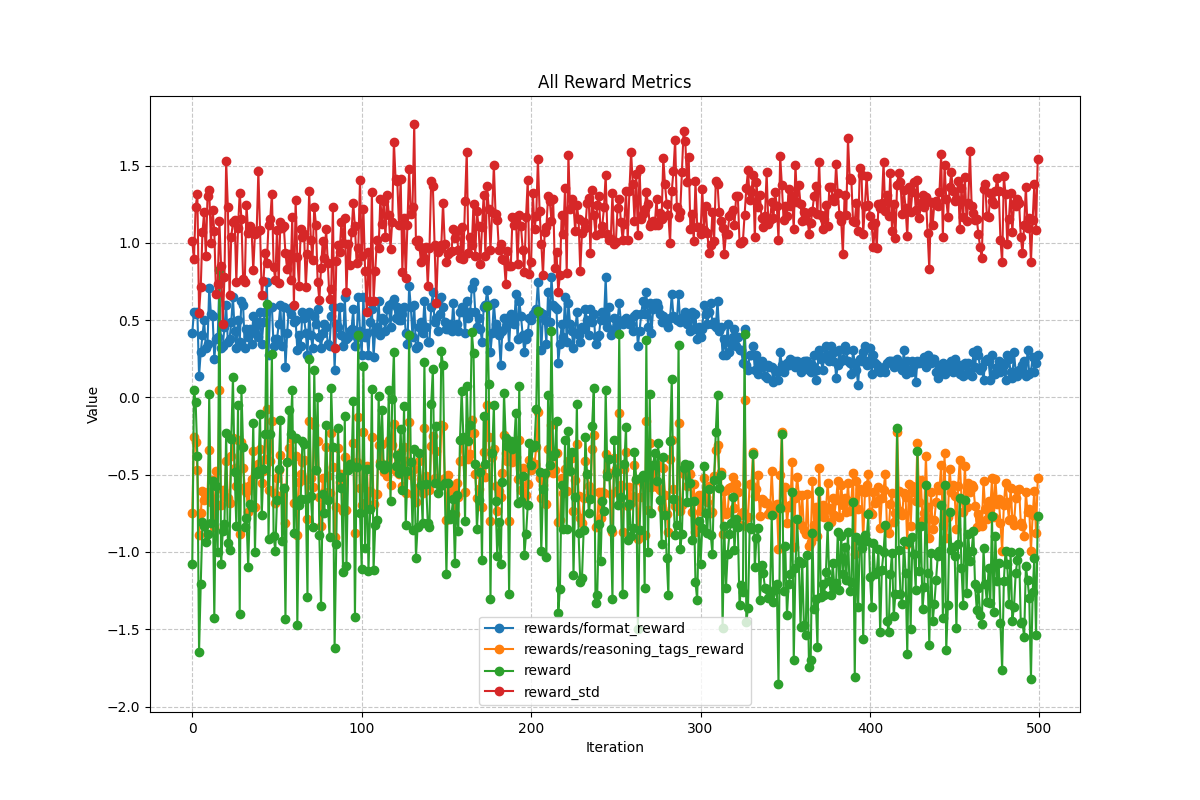}
    \caption{Reward curve without masking}
    \label{fig:wo_mask_reward}
\end{subfigure}

\begin{subfigure}[b]{0.495\textwidth}
    \centering
    \includegraphics[width=\linewidth]{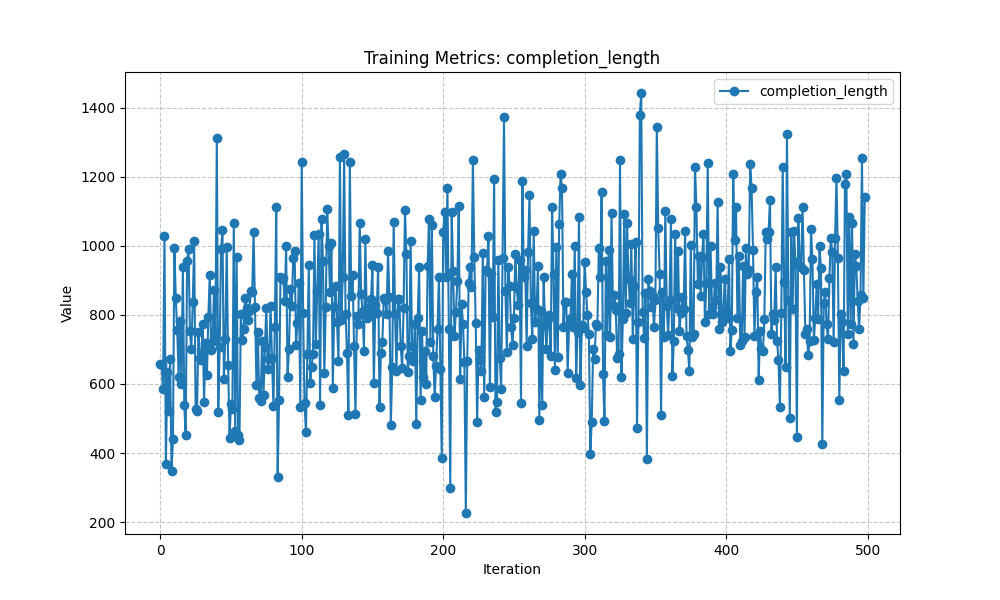}
    \caption{Response length curve with masking}
    \label{fig:w_mask_train}
\end{subfigure}
\hfill
\begin{subfigure}[b]{0.495\textwidth}
    \centering
    \includegraphics[width=\linewidth]{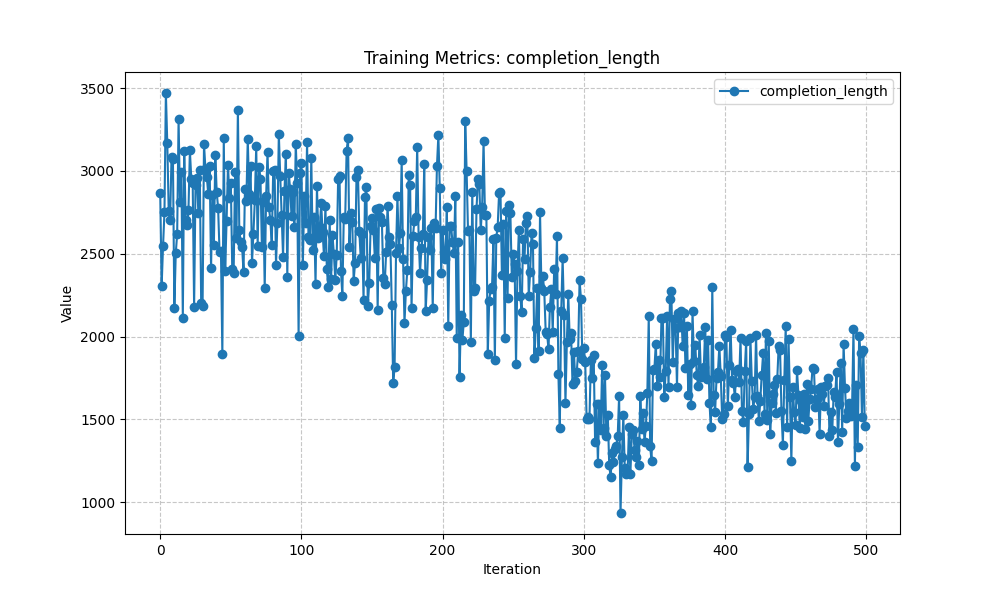}
    \caption{Response length curve without masking}
    \label{fig:wo_mask_train}
\end{subfigure}

\caption{Impact of Truncated Completion Masking on Training Stability}
\label{fig:masking_analysis}
\end{figure}

Our observations reveal that truncated long outputs can induce notable gradient fluctuations and result in unstable training processes (Figure~\ref{fig:masking_analysis}). To mitigate this issue, we mask the truncated completions to disregard their reward values and gradient updates. This approach effectively stabilizes optimization by omitting samples surpassing a predefined length limit.

\textbf{Tag Randomization for Robustness}

Inspired by DeepSeek-R1's reasoning completions, we introduce randomization in the order of reasoning tags in the prompt during training. Specifically, for each question, we retain the top 5 tags and randomly sample 0--5 additional tags from the remaining set, shuffling their order in the prompt. This approach reduces overfitting to fixed reasoning patterns and encourages the model to generalize reasoning strategies.

\textbf{Supervised Fine-Tuning (SFT)}

We conducted supervised fine-tuning using the DeepSeek-R1-Distill-Qwen-1.5B model as our base architecture. The model was trained for 5 epochs with a learning rate of 1e-5 using a cosine scheduler with no minimum learning rate. We applied a weight decay of 1e-4 for regularization. The training was performed on a single NVIDIA A6000 GPU with a micro-batch size of 1 and no gradient accumulation. We used a context window of 4096 tokens and trained on the "S1 Structured 500 Completions" dataset. The model was trained using mixed precision (bfloat16) with fully sharded data parallelism for memory efficiency. The final model was saved to the "SR-SFT" directory without pushing to the Hugging Face Hub.

\textbf{Reinforcement Learning (RL)}

For the reinforcement learning phase, we used our fine-tuned SR-SFT model as the starting point. We implemented GRPO (Generalized Reinforcement from Policy Optimization) training on the DeepScaleR-Preview-Dataset. The model was trained with bfloat16 precision with Flash Attention 2 for improved efficiency. We used VLLM for inference with 70\% GPU memory utilization and a maximum sequence length of 4096 tokens. Training was configured with a batch size of 6 per device, gradient accumulation of 4 steps, and gradient checkpointing to manage memory usage. We used a learning rate of 1e-6 with a cosine scheduler having a minimum learning rate at 10\% of the initial value. For each training example, we generated 6 completions with a temperature of 0.6. The reward function combines format scoring and max flow scoring (fixed weights 1.0 and 2.0). Format scoring serves as a fundamental component integrated into all reward functions. The training was conducted on 4 NVIDIA A100 GPUs. The final model was saved to the "SR-FLOW".

\subsection{Structured reasoning through Fill In the Middle API}
\label{reasoning prompt}
The full message template is structured the same as the prompt in Appendix~\ref{full_prompts}.

The API call is implemented as:
\begin{verbatim}
messages = [
    {"role": "user", "content": full_message},
    {
        "role": "assistant",
        "reasoning_content": "<rephrase>\nOkay, I will organize my thoughts 
        process in a hierarchical manner.\n</rephrase>\n<",
        "content": "",
        "prefix": True
    }
]
response = await self.client.chat.completions.create(
    model=model,
    messages=messages
)
\end{verbatim}

In particular, we initialize the reasoning process by injecting a <rephrase> tag and a metacognitive statement. This approach is effective in guiding DeepSeek-R1 to perform structured reasoning in a zero-shot setting, leading to more stable and organized reasoning patterns without additional training.

\newpage
\subsection{Limitation}
\label{limitation}
While our structured approach has demonstrated effectiveness across mathematical reasoning, natural sciences, and logical reasoning tasks, its impact on non-reasoning tasks such as story generation and agent API calls remains unclear. Although our MAX-Flow algorithm provides a more effective way than Perplexity to evaluate the importance of reasoning steps for answers, it requires researchers to process and implement model attention mechanisms more meticulously to obtain the reasoning step matrix. The training process also slightly increases memory overhead. These functionalities are not yet optimized by existing large model libraries and inference acceleration frameworks.

\subsection{Broader Impact}
\label{broader_impact}
\paragraph{Positive Impact}
Our research further validates the effectiveness of high-quality data fine-tuning combined with reinforcement learning training, providing examples for research in other tasks. We discovered that LLMs, similar to humans in cognitive science, can improve reasoning effectiveness through structured reasoning. By encouraging spontaneous labeling of structured reasoning processes, we can simplify reasoning while maintaining performance. This raises several considerations: should we develop deeper reward-based reasoning processes? Are there more complex reasoning structures (like CoT, XML, or code formats) that could benefit models across multiple tasks? Additionally, our structured reasoning approach can be applied to early exit strategies, test-time scaling, or Monte Carlo tree search phases, benefiting other tasks. Our further analysis reveals that different model layers show varying attention spans to reasoning steps - some layers primarily focus on the immediate previous step, while others attend to a broader range of steps. This insight opens possibilities for customized model optimization, model KV cache reduction, and inference pruning optimization (we found that reducing cache for layers focusing only on the previous step doesn't significantly impact performance, while the opposite can lead to question forgetting). Furthermore, structured reasoning enables clearer definition of reasoning steps, rather than relying on logical transition words like "but" or line breaks for parsing, making it more conducive to reasoning compression and analysis tasks.

\paragraph{Negative Impact}
Our method for evaluating reasoning effectiveness involves injecting irrelevant or harmful reasoning steps for accurate assessment. However, this approach could potentially be misused for reasoning attacks or harmful reasoning bias manipulation. Through layer-wise analysis, we discovered different attention interests across layers, which might make models vulnerable to attacks based on parameter modifications that simply redirect model attention patterns. This could lead to significant performance degradation, resulting in issues like forgetfulness (forgetting the input question after reasoning for a while) or repetitive outputs (entering text output loops that cannot exit, causing substantial computational resource waste). These issues require further attention.

\end{document}